\documentclass[acmlarge,nonacm]{acmart}
\usepackage{tabularx}
\usepackage{array}
\usepackage{gensymb}
\usepackage{booktabs}
\usepackage{xcolor}
\usepackage{graphicx}
\usepackage{tikz}
\usepackage{pgfplots}
\usepackage{forest}
\usepackage{harveyballs}
\usetikzlibrary{trees,positioning,shapes,shadows,arrows.meta}
\usepackage{wasysym}
\usepackage{colortbl}
\usepackage{subcaption}
\usetikzlibrary{positioning}
\usepackage{caption}
\usepackage{algorithm}
\usepackage[utf8]{inputenc}
\usepackage{algpseudocode}
\usepackage{amsmath}
\usepackage{amsthm}
\usepackage{amsfonts}
\usepackage{multirow}

\usepackage{xcolor}
\usepackage{graphicx}
\definecolor{myorange}{HTML}{F7941D}
\definecolor{myblue}{HTML}{27AAE1}  
\definecolor{mygreen}{HTML}{39B54A} 
\definecolor{constrast_steering_green}{HTML}{2f9e44}
\definecolor{constrast_steering_red}{HTML}{e03131}      
\definecolor{constrast_steering_blue}{HTML}{1970c2}

\newcommand{\green}{\textcolor{green}{\CIRCLE}}
\newcommand{\yellow}{\textcolor{yellow}{\CIRCLE}}
\newcommand{\red}{\textcolor{red}{\CIRCLE}}

\AtBeginDocument{%
  }
\setcopyright{acmlicensed}
\copyrightyear{2025}
\acmYear{2025}
\acmDOI{XXXXXXX.XXXXXXX}
\acmJournal{CSUR}
\acmVolume{1}
\acmNumber{1}
\acmArticle{1}
\acmMonth{1}

\definecolor{linearcontrastive}{RGB}{200,200,255} 
\definecolor{affinesteering}{RGB}{200,255,200} 
\definecolor{dynamicstrength}{RGB}{255,200,200} 
\definecolor{multiplemodel}{RGB}{200,255,255} 
\definecolor{finetuningalternatives}{RGB}{255,200,255} 
\definecolor{sparseautoencoders}{RGB}{255,255,200} 

\begin{document}
\title{Representation Engineering for Large-Language Models: Survey and Research Challenges}

\author{Lukasz Bartoszcze}
\email{lukasz.bartoszcze@wisent.ai}
\affiliation{%
  \institution{Wisent AI}
  \city{San Francisco}
  \country{United States}
}

\affiliation{%
  \institution{University of Warwick}
  \city{Coventry}
  \country{United Kingdom}
}

\author{Sarthak Munshi}\thanks{This research was conducted independently and does not reflect the views of Amazon Web Services, Inc. or Perplexity AI, Inc.}
\email{sarthakmunshi@gmail.com}
\affiliation{%
  \institution{Amazon Web Services}
  \city{Seattle}
  \country{United States}
}
\author{Bryan Sukidi}
\email{bryan.sukidi@gmail.com}
\affiliation{%
  \institution{University of North Carolina at Chapel Hill}
  \city{Chapel Hill}
  \country{United States}
}
\author{Jennifer Yen}
\email{jennifer.yenjy@gmail.com}
\affiliation{%
  \institution{Perplexity}
  \city{San Francisco}
  \country{United States}
}
\author{Zejia Yang}
\email{zy369@cam.ac.uk}
\affiliation{%
  \institution{University of Cambridge}
  \city{Cambridge}
  \country{United Kingdom}
}
\author{David Williams-King}
\email{d.williams-king@mila.quebec}
\affiliation{%
  \institution{Mila}
  \city{Montreal}
  \country{Canada}
}

\author{Linh Le}
\email{linh.le@uts.edu.au}
\affiliation{%
  \institution{University of Technology Sydney}
  \city{Sydney}
  \country{Australia}
}

\author{Kosi Asuzu}
\email{asuzukosiie@gmail.com}
\affiliation{%
  \institution{Wisent AI}
  \city{San Francisco}
  \country{United States}
}


\author{Carsten Maple}
\email{cm@warwick.ac.uk}
\affiliation{%
  \institution{University of Warwick}
  \city{Coventry}
  \country{United Kingdom}
}

\begin{abstract}

Large-language models are capable of completing a variety of tasks, but remain unpredictable and intractable. Representation engineering seeks to resolve this problem through a new approach utilizing samples of contrasting inputs to detect and edit high-level representations of concepts such as honesty, harmfulness or power-seeking. We formalize the goals and methods of representation engineering to present a cohesive picture of work in this emerging discipline. We compare it with alternative approaches, such as mechanistic interpretability, prompt-engineering and fine-tuning. We outline risks such as performance decrease, compute time increases and steerability issues. We present a clear agenda for future research to build predictable, dynamic, safe and personalizable LLMs.

\end{abstract}

\ccsdesc[500]{Computing methodologies~Neural networks}
\ccsdesc[500]{Computing methodologies~Machine learning approaches}
\ccsdesc[500]{Computing methodologies~Machine learning algorithms}

\keywords{representation engineering, neural networks, activation steering, model editing, activation analysis, linear probing, large-language models, machine learning, artificial intelligence}

\maketitle

\section{Introduction}

Large Language Models (LLMs) have powered a unique breakthrough in machine learning capabilities by showing how new capabilities emerge with scale
\cite{wei2022emergent, bahri24}. Using large volumes of data and unprecedented computational resources, these models have demonstrated capabilities generalizing across a wide range of tasks and surpassing both human and previous, narrow AI system performance \cite{li2023survey, liu2023recent}.

LLMs now consistently outperform previous approaches across standard language understanding and reasoning benchmarks, showing particular strength in tasks requiring complex logical reasoning and multi-step problem solving \cite{li2024comprehensive, yang2023harnessing}. For example, in healthcare, LLMs demonstrate the ability to analyze complex medical cases, suggest potential diagnoses, identify drug interactions, and assist in treatment planning \cite{liu2023survey}. LLMs have also had a profound impact on software development, and are able to understand programming concepts, generate functional code, and debug complex programs \cite{chen2023survey}, dramatically accelerating the development process. In scientific research, LLMs are capable of processing vast amounts of scientific literature, identifying patterns across disparate fields, and suggesting novel research directions \cite{dai2023language}. These models also perform well in education by providing personalized tutoring, adapting to individual learning styles, and explaining complex concepts in a way personalized to the student's understanding \cite{holmes2022state}. Across law and finance, the models have demonstrated sophisticated comprehension of complex documents, regulatory requirements, and contractual terms \cite{lin2019artificial}. 

Perhaps most importantly, these models show an advanced ability to transfer knowledge between domains \cite{dai2023language}, combining insights  from different fields to solve novel problems they have not seen in the training data \cite{wei2022emergent}. Their reasoning capabilities can be extended by breaking down complex problems into manageable steps \cite{wei2023chain}, mirroring human problem-solving approaches. Through this, intelligence becomes a scale problem, and under the right circumstances, LLMs are able to equally and eventually surpass human intelligence. While challenges remain in ensuring consistent factual accuracy \cite{qin2023survey} and adequate assurance to guarantee trustworthiness \cite{zhao2023survey}, these models represent a transformative shift in how human operators utilise technology. Their ability to understand context, maintain coherent long-term reasoning, and adapt to new tasks shows their potential for assisting, automating and substituting most areas of human work. 

Much of the improvement in LLM capabilities can be attributed to their size. The models have been increasing and improving consistently over the years, from BERT's 340M parameters \cite{devlin2019bert} to GPT-3's 175B \cite{brown2020language} and Llama's 3.1. 405B parameters \cite{dubey2024llama}. While this increase in scale has significantly enhanced their performance, it has also introduced challenges. The sheer size and complexity of state-of-the-art models makes them difficult to verifiably control and modify \cite{bender21}. With parameter sizes in billions, models often operate as black-box entities, with researchers having little oversight over the actual interactions happening in the hidden layers of the model. In addition to scaling, an emerging trend in LLMs is their improved reasoning ability. This advancement does not strictly depend on scaling but rather on architectural innovations, training techniques, and the ability to generalize across tasks \cite{deepseekai2024deepseekv3technicalreport}. Reasoning capabilities enable LLMs to perform complex problem-solving, logical inference, and even exhibit some degree of common-sense understanding. However, this added sophistication further complicates efforts to ensure transparency and accountability in these systems.
\subsection{Representation Engineering}
\begin{figure}[h]
  \centering
  \includegraphics[width=0.45\linewidth]{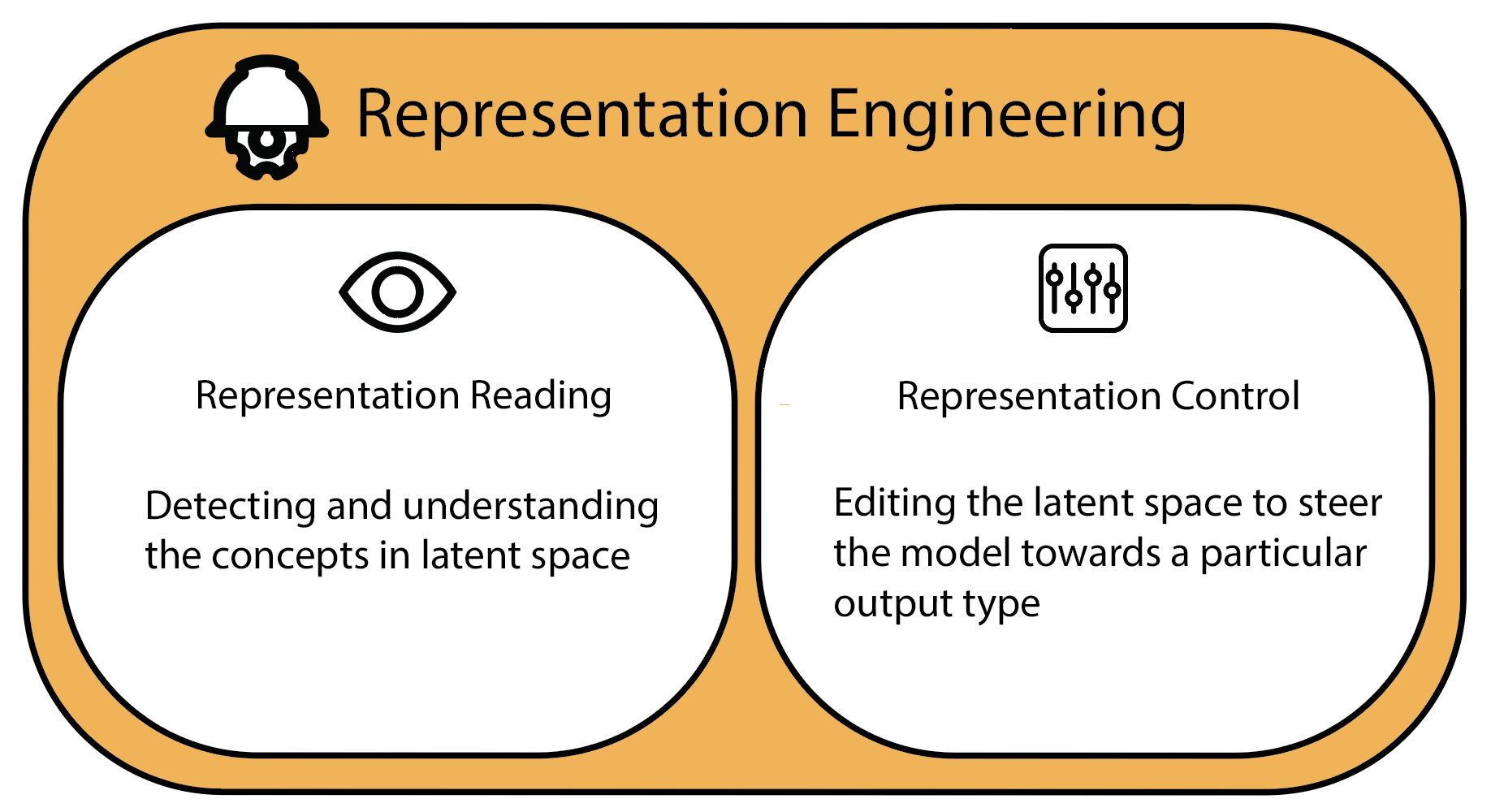}
  \caption{Representation Engineering: Representation Reading and Representation Control.}
  \Description{A diagram explaining Representation Engineering, divided into two sections: Representation Reading (detecting concepts in latent space) and Representation Control (editing latent space to steer model outputs).}
\end{figure}
Several approaches to the explainability of LLM have been applied over the years, including mechanistic interpretability \cite{rai24} \cite{zhao24exppersp} \cite{bereska24}, latent saliency maps \cite{schroder23} \cite{brocki19}, attribution-based methods \cite{nielsen22} \cite{ancona19}, counterfactual analysis \cite{cheng24} \cite{youssef24}, and probing techniques \cite{dong23} \cite{wang2024probing}. Representation engineering takes a different approach: global identification of high-level concepts by stimulating the network with contrasting pairs of inputs to identify differences in concepts and extract related features \cite{zhao24survey}. Instead of attempting to deconstruct the network into individual units, it places the activation across a global population of neurons at the heart of the analysis.

Generalizations in the form of representations are important because they model how human brains work, creating abstractions from information, rather than merely retrieving it  \cite{he2024multilevel} \cite{blank2023large}. Creating accurate representations of high-level concepts is a prerequisite for accurate out-of-distribution performance and emergence of AI thought and intelligence \cite{ferry2023emergence} \cite{yetman2025representation} \cite{mahowald2024dissociating} \cite{harding2023operationalising} \cite{goldstein2024does}. Symbolic data structures in the form of representations are the determining factor from neuroscientific perspective of whether the model merely encodes information like a "stochastic parrot" or whether it operates like a human brain \cite{blank2023large} \cite{chalmers2025propositional} \cite{yetman2025representation} \cite{hardy2023large} \cite{goldstein2024does}. 

Representation Engineering is performed in two distinct steps. The first one, Representation Reading, focuses on locating and extracting representations from the network, while the second one, Representation Control, aims to use this knowledge to steer the model towards a particular outcome by modifying the internal representations~\cite{li2024inference}.

\subsection{Related Work}

\paragraph{Existing Survey Work}

Existing deep learning literature produced a number of studies presenting large-scale overview of techniques and open problems in AI explainability (XAI) including general AI explainability \cite{gohel2021explainable}, deep learning explainability \cite{csahin2024unlocking}, black box models  \cite{choudhary2022interpretation}, large-language model XAI \cite{zhao24survey} \cite{danilevsky2020survey} \cite{ehsan2024explainability} \cite{wang2024knowledge}, neural network concept explanation \cite{lee2023neural} \cite{rauker2023toward} or medical XAI \cite{sheu2022survey}. More narrow studies focus on mechanistic interpretability \cite{bereska24} \cite{ferrando24} \cite{kastner24}, LLM knowledge encoding \cite{wang2024knowledge}, comparing models on the representation level \cite{klabunde2023similarity} or probing \cite{belinkov2022probing}.

In that, some surveys provide a brief overview of representation engineering as a counterpoint to their main focus. Zhao et al. \cite{zhao24exppersp} provides a landscape survey for modern explainability with a brief overview of representation engineering and analyzes representation engineering in relation to mechanistic interpretability. Representation engineering is also briefly analyzed as a potential alternative to existing explainability techniques in \cite{zhao24survey}. This study is fundamentally different. It is the first study to review the work on representation engineering, an emerging field with high empirical validation for its techniques. It aims to highlight and systematize the techniques in this growing field to provide insights necessary for the creation of stable, general-purpose reading and interventions that can be applied across all use cases with a top-down interpretation. 

\paragraph{Latent Saliency Maps (LSMs)}  

Latent Saliency Maps show how internal representations influence predictions in language models by highlighting relevant activations, as demonstrated in emergent world models in sequence tasks \cite{li2022emergent}. An extension of general Latent Saliency Maps, Concept Saliency Maps (CSMs) identify high-level concepts by calculating gradients of concept scores \cite{brocki2019concept}. 

\paragraph{Concept Bottleneck Models (CBMs)} 

Pre-LLMs, Concept Bottleneck Models (CBMs) have been created as a deep learning architecture that has an intermediate layer that forces models to represent information through human-understandable concepts, enabling interpretability and direct intervention \cite{koh2020concept}. CBMs have been extended to Concept Bottleneck Generative Models (CBGMs), where a dedicated bottleneck layer encodes structured concepts, preserving generation quality across architectures like GANs, VAEs, and diffusion models \cite{ismail2023concept}. However, CBMs suffer from "concept leakage," where models bypass the bottleneck to encode task-relevant information in uninterpretable ways, which can be mitigated using orthogonality constraints and disentangled concept embeddings \cite{mahinpei2021promises, ismail2023concept}. Concept Bottleneck Large-Language Models (CB-LLMs) integrate CB layers into transformers, demonstrating that interpretable neurons can improve text classification and enable controllable text generation by modulating concept activations \cite{sun2024concept}. CBMs tie inference to a specific "concept" (representation), but usually have lower accuracy than concept-free alternatives. Their effectiveness depends on the completeness and accuracy of the process of identifying the concept, leading to new generations of models that perform automated concept discovery \cite{zarlenga2022concept, ismail2023concept, kim24}. 

\paragraph{Concept Activation Vectors (CAVs)}  

Concept Activation Vectors (CAVs) are numerical representations of concepts across layers. They provide a way to probe learned representations in neural networks by identifying directions in latent space that correspond to human-interpretable concepts \cite{nicolson24explain}. However, they are not stable across different layers of a model but evolve throughout the network \cite{nicolson24explain}. The entanglement of multiple concepts within a single CAV makes it difficult to assign the meaning to learned representations \cite{nicolson24explain, schmalwasser24}. Concept Activation Regions (CARs) enhance concept-based explanations by generalizing Concept Activation Vectors (CAVs) to account for non-linear separability, leading to more accurate global explanations \cite{crabbe2022concept}.  

Although directly tied to representation engineering, LSMs, CBMs and CAVs are a method primarily used for deep learning models \cite{desantis24, schmalwasser24} and have not been extensively applied to large-language models.

\subsection{Theoretical Foundations of Representation Engineering}

\subsubsection{Linear Representation Hypothesis}

The theory of why representation engineering can be used effectively is based on the Linear Representation Hypothesis (LRH). LRH postulates that high-level concepts and functions are encoded in the activations of neural networks as linear or near-linear features, identified as directions or subspace in the latent space of the model. \cite{park2023linear, mikolov13}. 

Representation engineering is built upon linear representation hypothesis, as it relies on the ability to isolate, manipulate, and interpret specific features within a model’s latent representation space using linear methods. If common features like sentiment, gender, or style were not encoded in approximately linear subspaces, techniques such as vector arithmetic, projection, and basis decomposition would not work effectively. The success of interventions like linear probes \cite{alain16}, activation steering, bias removal, and feature control hinges on the assumption that these properties can be modified independently without complex non-linear entanglements. Linear representations in LLMs emerge naturally from the interaction of the next-token prediction objective and the implicit bias of gradient descent, rather than from architectural constraints \cite{jiang2024origins}. If representations were highly non-linear or span multiple concepts, changing one attribute might unpredictably alter others, making controlled modifications impossible. Thus, the practical tools of representation engineering—feature extraction, interpretability, and controlled model editing—are fundamentally dependent on representations being structured in a way that allows for linear operations to meaningfully adjust model behavior.

The LRH is supported by extensive empirical findings across NLP and vision models, demonstrating that high-level features are often encoded as linear directions in activation space. Early evidence comes from word embeddings such as Word2Vec, where simple vector arithmetic (e.g., "king" - "man" + "woman" = "queen") suggested that semantic attributes are represented in a linear fashion \cite{mikolov13}. This idea extends to modern transformer-based models, where probing studies have shown that features like syntax trees \cite{hewitt19}, part-of-speech tags \cite{tenney19}, and named entities \cite{liu19named} can be recovered using linear classifiers with high accuracy. Large-language models have been shown to be effectively probed for space and time \cite{gurnee23}. Similarly, truthfulness direction can also be found to significantly improve model performance on hallucination benchmarks \cite{li2024inference}. In factual recall tasks, Meng et al. \cite{meng22} demonstrated that specific residual stream activations encode factual associations in GPT models, with direct interventions shifting model predictions toward or away from a known fact. Burns et al. \cite{burns2023discovering} found that unsupervised contrastive probes can identify “truth” directions in LLM activations, though generalization issues persist \cite{levinstein2024still}. There is also evidence that individual concepts are in fact vectors \cite{piantadosi2024concepts}. In adversarial robustness literature, the presence of universal jailbreaks is highly predicated on the idea of LHR being true \cite{moosavi2017universal}.

Bau et al. \cite{bau2020rewriting} showed that individual neurons and linear directions in CNNs correspond to human-interpretable object concepts, supporting the idea that representations align with linear subspaces rather than single neurons. These results collectively reinforce that while not all representations are perfectly linear \cite{elhage2022toy}, many high-level abstractions in deep networks behave as approximately linear features, forming the basis for practical representation engineering. If LRH is not true, representation engineering techniques like activation steering, bias removal, and linear probes may be unreliable, as modifying one attribute could unintentionally alter others, limiting precise control over model behavior and leading to lack of generalizable hyperparameters that can be set for multiple problems.

Even though strong evidence supporting LHR being true exists, it is still a hypothesis and not a fully proven, absolute fact. In fact, probing shows that more complex internal representations are highly non-linear, encoding as detailed information as the board state during a game \cite{li2022emergent}. Linear representations may not capture the real structure but instead emerge during pre-training as bounds on the model's cognitive capacity \cite{yan2024exploring}. In particular, much stronger support exists for the Weak LRH (assuming that some features are linear) than for Strong LRH (assuming that all features are linear). 

Studies of the validity of LRH \cite{park24} provide a formal framework linking different interpretations of linear representation—subspace structure, probing, and intervention—demonstrating their fundamental connections. They introduce the concept of a causal inner product, refining how linear directions in representation space should be identified to ensure meaningful interventions. These findings support Weak LRH by confirming that many high-level features are encoded in linear subspaces, while also highlighting that naive geometric assumptions can lead to misleading conclusions.  Other studies demonstrate that simple linear probes can reliably identify and manipulate a "truth direction," reinforcing the idea that at least some high-level concepts are structured linearly in model representations \cite{marks24, kadavath2022language}. This strengthens the theoretical basis of representation engineering, showing that while linear representations are exploitable, their structure may require more precise mathematical treatment to be reliably manipulated and, while performant in some aspects, cannot be reliably generalized for all cases.

While the empirical effectiveness of representation engineering is well-supported by evidence from other parts of the literature studying LRH, several studies show, that strong LRH may not be satisfied. Therefore, a generalized representation engineering intervention may not be possible and specific techniques need to be implemented to detect and mitigate nonlinear interactions.

\subsubsection{Superposition Hypothesis}

Superposition hypothesis states that the neural network represents more interpretable representations than there are dimensions in the representation space \cite{elhage2022toy}. Instead of dedicating one axis of the representation space per feature, the network represents features as distinct directions in a high-dimensional activation vector. This hypothesis supports the empirical result of why neural representations are often distributed and overlapping \cite{huang2023rigorously}; the model effectively simulates a larger set of features than the number of neurons by superimposing features.

If true, the superposition hypothesis presents validation to the representation engineering approach, but also a challenge in practical implementation and development of general methods for representation engineering. Because of how neurons are encoded, a top-down approach allows to mitigate neuron polysemanticity and extract the net result of activations in the latent space. On the other hand, an intervention on a particular subset of the latent space may lead to changes in other representations, leading to decrease in capabilities and overall coherence. Furthermore, stacking several interventions may result in further unintended consequences, because one intervention may boost some effect of the other or nullify it. 

\section{Representation Reading}

\begin{figure}[h]
  \centering
  \includegraphics[width=0.9\linewidth]{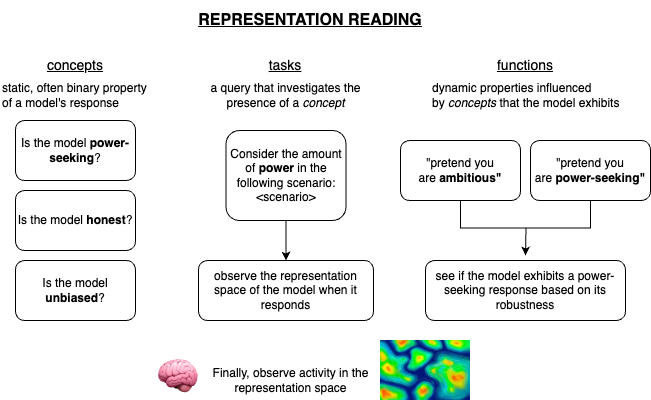}
  \caption{Entities in representation reading: concepts, tasks, functions.}
  \label{fig:rep_reading}
\end{figure}

Representation Reading seeks to extract information about what activations correspond to a particular representation from the neural network, thus showing how LLMs represent and process information. As shown in Figure~\ref{fig:rep_reading}, a representation can correspond to multiple entities, specifically a concept, task, or function:
\begin{itemize}
    \item A \textit{concept} is a static property of the model like truthfulness, harmlessness, or morality.
    \item A \textit{task} is a particular user query that for which a beneficial representation can be amplified to complete the task more effectively.
    \item A \textit{function} is a dynamic property of the model like outputting correct Python code, power-seeking or answering a question in Spanish.
\end{itemize}
This section outlines methods to identify such representations. Causal manipulation experiments reveal that altering these abstract representations leads to predictable changes in model output, confirming that the network uses them for decision-making rather than passively encoding correlations~\cite{ferry2023emergence}. Although representation engineering interventions require white box access, representation reading can be applied to black box models to predict overall model performance and detect harmful versions of LLMs \cite{sam2024eliciting}.

\subsection{Linear Artificial Tomography}

\begin{figure}[h]
  \centering
  \includegraphics[width=0.6\linewidth]{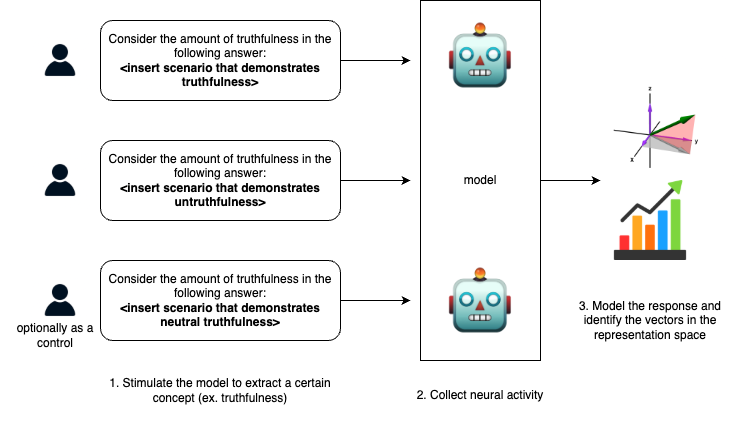}
  \caption{A high-level example of a LAT experiment.}
  \Description{}
\end{figure}

The overarching idea of representation reading is to use differing inputs to stimulate the neural activity of the model, and to use the differences in observed activity to predict which activations correspond to model behavior. A foundational technique to detect such activations is Linear Artificial Tomography (LAT)~\cite{zou2023representation}. 
LAT reading requires defining a stimulus and a task for detecting the neural activity corresponding to it. The stimulus can be as simple as the following prompt (with ``\_\_\_'' added to show where output would be generated):
\begin{quote}
Consider the amount of <CONCEPT> in the following: <STIMULUS>. The amount of <CONCEPT> is \_\_\_
\end{quote}
The neural activity of the network when stimulated with this model is then compared with the activity of the model when given either a) a contrasting stimuli or just b) a reference stimuli with no specific concept or function. Based on this data, a linear model is used to identify a direction to detect the direction of the identified representation. These models can either be supervised or unsupervised. In practical applications, either a supervised linear probe or an unsupervised Principal Component Analysis (PCA) is applied.

For a concept $c$, given a model $M$, a function $\text{Rep}$ that extracts representations, and a stimulus set $S$, the neural activity is collected as:
\begin{equation}
A_c = { \text{Rep}(M,T_c(s_i))[-1] \mid s_i \in S }
\end{equation}
where $T_c(s_i)$ represents the task template applied to stimulus $s_i$. To determine the principal representation direction, PCA is applied to the difference vectors of neural activations:
\begin{equation}
v = \text{PCA}( { A_c^{(i)} - A_c^{(j)} } )
\end{equation}
where $A_c^{(i)}$ and $A_c^{(j)}$ are pairs of neural activities from different stimuli. To manipulate the model's representation, a projection operation is performed:
\begin{equation}
R' = R - \frac{R^T v}{| v |^2} v
\end{equation}
where $R$ is the original representation and $R'$ is the transformed representation after adjusting along the reading vector $v$. For more details, see \citet{zou2023representation}.

\subsection{Probing}

\tikzset{
  basic/.style = {
    draw, rectangle, 
    font=\sffamily,
    align=center,
    rounded corners=2pt,
    thin
  },
  rootStyle/.style = {
    basic,
    fill=green!30
  },
  levelOne/.style = {
    basic,
    fill=red!30
  },
  levelTwo/.style = {
    basic,
    fill=blue!30
  },
  finalLeaf/.style = {
    basic,
    fill=orange!30
  },
  edge from parent/.style={
    draw=black, 
    edge from parent fork right
  }
}

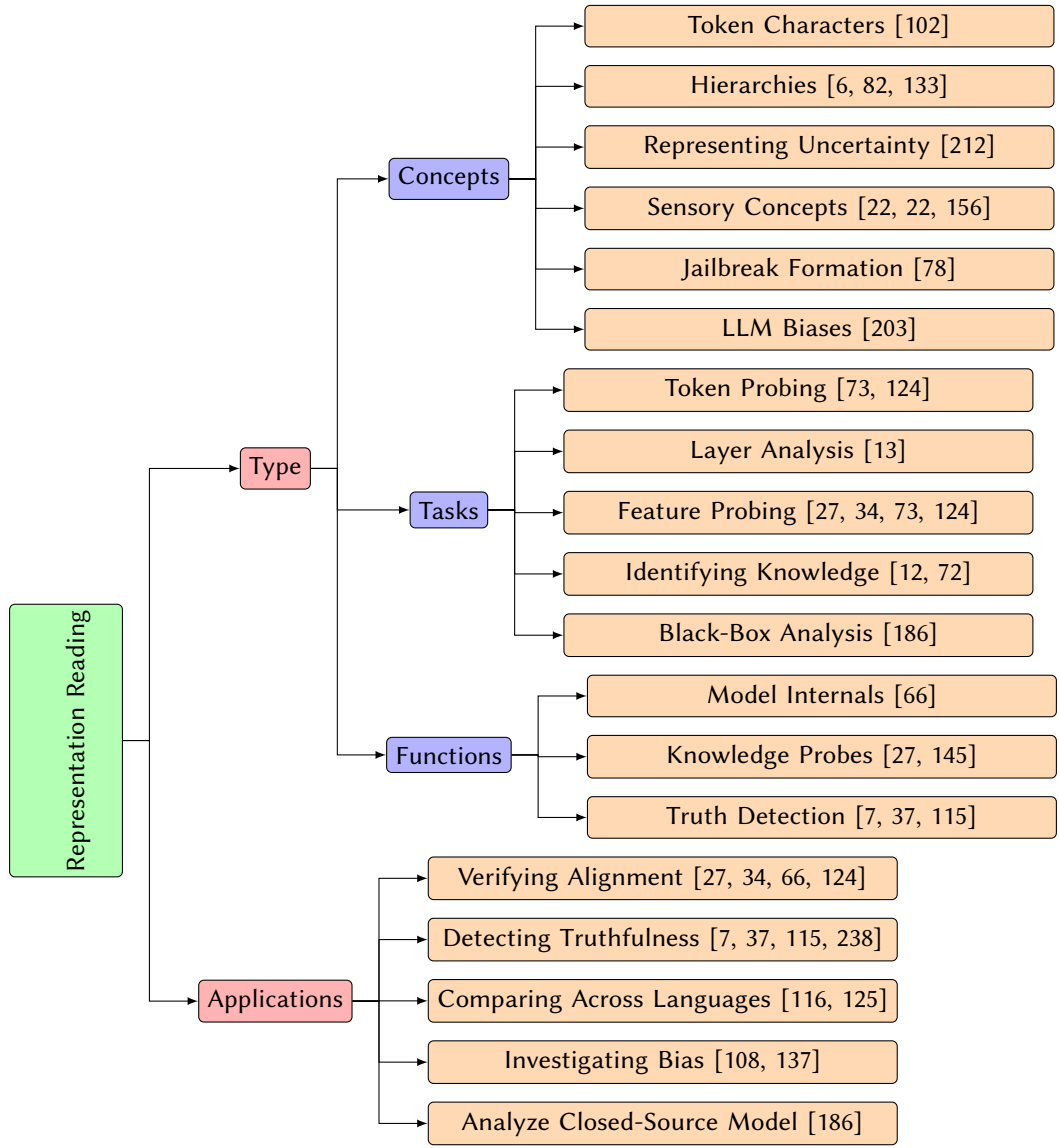
\begin{figure}
\centering

\tikzset{
  basic/.style = {
    draw, rectangle, 
    font=\sffamily,
    align=center,
    rounded corners=2pt,
    thin
  },
  rootStyle/.style = {
    basic,
    fill=green!30
  },
  levelOne/.style = {
    basic,
    fill=red!30
  },
  levelTwo/.style = {
    basic,
    fill=blue!30
  },
  finalLeaf/.style = {
    basic,
    fill=orange!30,
    text width=6cm,
  },
  edge from parent/.style={
    draw=black, 
    edge from parent fork right
  }
}

\begin{forest}
  for tree={
    grow=east,
    growth parent anchor=west,
    parent anchor=east,
    child anchor=west,
    l sep=10mm, 
    edge path={
      \noexpand\path[\forestoption{edge},->, >={latex}]
      (!u.parent anchor) -- +(10pt,0pt)
      |- (.child anchor)
      \forestoption{edge label};
    }
  }
  [ {\rotatebox{90}{Representation Reading}},  
    rootStyle,
    text width=0pt,
    minimum width=1.5cm,
    inner sep=2pt,
    anchor=center,
    l sep=10mm
    [Applications, levelOne, l sep=10mm
      [Analyze Closed-Source Model \cite{sam2025predicting}, finalLeaf]
      [Investigating Bias \cite{lu24, kotek2023gender}, finalLeaf]
      [Comparing Across Languages \cite{li2024vision, li24ranker}, finalLeaf]
      [Detecting Truthfulness \cite{yin2024characterizing, azaria2023internal, chen2024inside, levinstein2024still}, finalLeaf]
      [Verifying Alignment \cite{ghandeharioun2024patchscope, ch2023androids, lillms, bronzini2024unveiling}, finalLeaf]
    ]
    [Type, levelOne, l sep=10mm
      [Functions, levelTwo, l sep=10mm
        [Truth Detection \cite{levinstein2024still, chen2024inside, azaria2023internal}, finalLeaf]
        [Knowledge Probes \cite{manigrasso2024probing, bronzini2024unveiling}, finalLeaf]
        [Model Internals \cite{ghandeharioun2024patchscope}, finalLeaf]
      ]
      [Tasks, levelTwo, l sep=10mm
        [Black-Box Analysis \cite{sam2025predicting}, finalLeaf]
        [Identifying Knowledge \cite{belinkov2022probing, guo2024steering}, finalLeaf]
        [Feature Probing \cite{ch2023androids, lillms, bronzini2024unveiling, gurnee2023finding}, finalLeaf]
        [Layer Analysis \cite{belrose2023eliciting}, finalLeaf]
        [Token Probing \cite{lillms, gurnee2023finding}, finalLeaf]
      ]
      [Concepts, levelTwo, l sep=10mm
        [LLM Biases \cite{tang2023llamas}, finalLeaf]
        [Jailbreak Formation \cite{he2024jailbreaklens}, finalLeaf]
        [Sensory Concepts \cite{ngo2024language, boldsen2022interpreting, boldsen2022interpreting}, finalLeaf]
        [Representing Uncertainty \cite{viswanathan2025geometry}, finalLeaf]
        [Hierarchies \cite{arps2022probing, liu2024active, hernandez2022ast}, finalLeaf]
        [Token Characters \cite{kaushal2022tokens}, finalLeaf]
      ]
    ]
  ]
\end{forest}
\caption{Graph showing Representation Reading, categorized by applications and type of representation.}
\label{fig:representation_reading}
\end{figure}

Probing aims to relate specific features to activation patterns in neural networks by training supervised models to map activations to target variables~\cite{ivanova2021probing} and eventually representations.  
Representations can be detected by examining the emergence of abstract, low-dimensional manifolds that encode semantic features shared by different inputs \cite{ferry2023emergence}. The complexity of the probing classifier influences the results, with simpler probes often providing more interpretable insights. More complex probes may achieve higher accuracy but risk inferring features not actually used by the network. Causal analysis techniques involve interventions in the representations to assess the impact on the model's original performance, revealing whether certain features are genuinely utilized. The choice of datasets for both the original model and probing tasks significantly affects the outcomes, making it critical to choose the right dataset to probe on \cite{levinstein2024still, kumar2022probing}.

Figure~\ref{fig:representation_reading} shows many example goals that can be achieved through representation reading. More is discussed in Section~\ref{sec:representation_reading:probing:applications}. The figure also shows specific techniques that are covered in Section~\ref{sec:representation_reading:probing:linear} and Section~\ref{sec:representation_reading:probing:other}.

\subsubsection{Applications of Probing}
\label{sec:representation_reading:probing:applications}

Probes have been applied to investigate representations such as LLM biases \cite{tang2023llamas}, jailbreak formation mechanisms \cite{he2024jailbreaklens} and even token embeddings of senses like sounds~\cite{ngo2024language}, shapes and color~\cite{boldsen2022interpreting}. Probes have been shown to encode uncertainty \cite{viswanathan2025geometry}, hierarchical linguistic constituency \cite{liu2024active}, cross-lingual retrieval \cite{guo2024steering}, Abstract Syntax Trees (ASTs)~\cite{hernandez2022ast}, hierarchical syntax \cite{arps2022probing}, model knowledge \cite{belinkov2022probing} or characters in a token \cite{kaushal2022tokens}.
Probing can even be applied in biological neural networks~\cite{ivanova2021probing} and black-box models \cite{sam2025predicting}.


Probes have been applied to check whether a model is acting against its maker's wishes \cite{liu24cognitive}, showing that models intentionally generate falsehoods, explaining hallucination and model misalignment via distinct prediction mechanisms. 
Several studies find effective solutions to detect truthfulness representations through different methods of probing \cite{levinstein2024still, ch2023androids, chen2024inside, yin2024characterizing, azaria2023internal}. 
Probes have also shown to be effective at identifying LLM performance across written languages using cosine similarity \cite{li24ranker}, revealing how activations for high-resource and low-resource languages differ. Through probing, it has been found that as language models scale, their vector spaces increasingly resemble the rich vector spaces of vision models \cite{li2024vision}. Probes have shown to be effective at detecting representations the model uses for distinguishing between its own knowledge and actions and those of others, supporting nuanced social reasoning tasks \cite{zhu2024language}. Probes have also been efficently applied to investigate societal biases \cite{lu24}. Despite Reinforcement Learning from Human Feedback (RLHF), the model retains gender bias \cite{kotek2023gender} while outright refusing to engage with race and religion related prompts. Manipulating internal belief representations leads to consistent changes in Theory of Mind performance, confirming that the model actively uses them for decision-making rather than merely storing passive correlations \cite{zhu2024language}.

\subsubsection{Linear Probes}
\label{sec:representation_reading:probing:linear}

Probing is often implemented through linear probes, simple linear classifiers trained on frozen features extracted from specific layers of a pre-trained neural network at inference time. By taking activations from a given layer as input and training to predict a specific property, such as part-of-speech tagging or sentiment analysis, linear probes show how well different layers capture certain types of information~\cite{alain16, kadavath2022language}. Given a hidden representation \( H_k \) at layer \( k \), a linear probe applies a weight matrix \( W \) and bias \( b \) followed by a softmax function:

\[ f_k(H_k) = \text{softmax}(W H_k + b) \]
where the softmax function is defined as:

\[ \text{softmax}(z_i) = \frac{e^{z_i}}{\sum_{j=1}^{D} e^{z_j}} \]
This transformation maps the hidden representation into a probability distribution over \( D \) classes.
The parameters of the linear probe \( W \) and \( b \) are trained using the cross-entropy loss:

\[ L = -\sum_{i=1}^{D} y_i \log(\hat{y}_i) \]
where \( y_i \) represents the true class label, and \( \hat{y}_i \) is the predicted probability for class \( i \). The objective is to minimize \( L \), thereby aligning the learned representations with the desired classification task.

Linear probes allow us to localise where specific types of information are encoded. These abstractions can be identified by observing the convergence of embeddings for tokens with similar semantic roles, even when they are perceptually distinct \cite{ferry2023emergence}. For example, truthfulness information tends to concentrate within ``exact answer tokens''~\cite{orgad2024llmsknowshowintrinsic}. Error detection methods can be significantly enhanced by focusing on these tokens. Intermediate representations of LLMs can be used to predict the types of errors they might make \cite{kadavath2022language}. 

Other methods add new components to this setup for more precise representation identification. 
Representational Similarity Analysis (RSA) uses pairwise cosine similarities between embeddings to reveal latent structure in task representations, enabling classification of task components without parameterized probes \cite{boldsen2022interpreting, yousefi2023context}. Contrast-Consistent Search (CCS) identifies truth representations by probing embeddings associated with the final token of a statement, aiming to extract subjective probabilities of truthfulness using supervised learning on labeled datasets \cite{levinstein2024still}.

\subsubsection{Other Types of Probes}
\label{sec:representation_reading:probing:other}

Logic Tensor Probes (LTP) train two-layer neural models to probe for specific predicates, treating each predicate's embedding method as a hyperparameter and optimizing probe depth for improved interpretability \cite{lillms}. LTPs are implemented by repurposing Logic Tensor Networks as shallow classifiers that assess logical consistency in frozen language model representations by encoding first-order logic constraints \cite{manigrasso2024probing}.

Probing Classifiers utilize transformer hidden states to predict hallucinations at token and response levels, leveraging layer-specific features for improved classification accuracy \cite{ch2023androids}. Probing Classifiers in \citet{bronzini2024unveiling} extract factual knowledge from LLM hidden states via activation patching, using classifiers to map token activations to logical predicates for claim verification.

INSIDE detects hallucinations by measuring semantic consistency in sentence embeddings using EigenScore, a covariance matrix determinant-based metric that captures dense semantic information \cite{chen2024inside}.
SAPLMA trains a classifier on LLM hidden states to predict the truthfulness of statements, focusing on out-of-distribution generalization and improving over probability-based methods \cite{azaria2023internal}. Pathoscopes uses the models’ own generation capabilities to explain their internal computations in natural language \cite{ghandeharioun2024patchscope}. Tuned lens target affine transformations that align intermediate hidden states with final layer logits \cite{belrose2023eliciting}. 
Intrinsic Dimensions correlates the geometric structure of token embeddings with classification loss, showing that tokens in high-loss prompts exist in higher-dimensional spaces \cite{viswanathan2025geometry}.

\section{Representation Control}

\begin{figure}[h]
  \centering
  \includegraphics[width=0.65\linewidth]{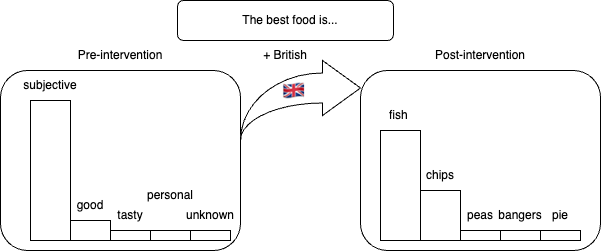} 
  \caption{Explanatory effect of an intervention on next token probabilities.}
  \label{fig:intervention}
  \Description{}
\end{figure}

Representation Control refers to the process of modifying or guiding the internal representations of concepts and functions identified in the Representation Reading step towards a particular outcome. Representation Control interventions primarily take the form of an insertion of a vector (or a set of vectors) between the layers of the model at inference without retraining the model. See Figure~\ref{fig:intervention} for an example.

\begin{figure}
\centering

\tikzset{
  basic/.style = {
    draw, rectangle, 
    font=\sffamily,
    align=center,
    rounded corners=2pt,
    thin
  },
  rootStyle/.style = {
    basic,
    fill=green!30
  },
  levelOne/.style = {
    basic,
    fill=red!30
  },
  levelTwo/.style = {
    basic,
    fill=blue!30
  },
  finalLeaf/.style = {
    basic,
    fill=orange!30,
    text width=8cm,
  },
  edge from parent/.style={
    draw=black, 
    edge from parent fork right
  }
}

\begin{forest}
  for tree={
    grow=east,
    growth parent anchor=west,
    parent anchor=east,
    child anchor=west,
    l sep=10mm, 
    edge path={
      \noexpand\path[\forestoption{edge},->, >={latex}]
      (!u.parent anchor) -- +(10pt,0pt)
      |- (.child anchor)
      \forestoption{edge label};
    }
  }
  [ {\rotatebox{90}{Representation Control}},  
    rootStyle,
    text width=0pt,
    minimum width=1.5cm,
    inner sep=2pt,
    anchor=center,
    l sep=10mm
    [Goal, levelOne, l sep=10mm
      [Truthfulness \cite{wang24steering, cao24steering, panickssery24, pham24, li2024inference, garg24, zhang2024truthx, choi24, rahn24, chen24, xiao24}, finalLeaf]
      [Performance \cite{panickssery24, zou2023representation, hendel2023context, li2023transformers, yang2025task, zhuang2024vector, ilharco2022editing, bowen2024beyond, pham2024robust, panigrahi2023task, devin2019plan, huang2024multimodal, luo2024task, scalena24, stoher24, bronzini2024unveiling, luo24, chalnev24, zhao24, ferrando24halluc, liu2024aligning}, finalLeaf]
      [Security \cite{bhattacharjee24, cao24steering, lee24, mattturner24, jorgensen24, luo24, pham24, li2024inference, dong24, garg24, feng24, choi24, kong24, zou24circuit, zhang24, postmus24, bronzini2024unveiling}, finalLeaf]
      [Personalization \cite{mattturner24, jorgensen24, subramani22, cao24steering, zou2023representation, liu2024aligning, wu24, chalnev24}, finalLeaf]
    ] 
    [Method, levelOne, l sep=10mm
      [Layer, levelTwo
        [Singular \cite{panickssery24, bhattacharjee24, mattturner24, jorgensen24, luo24}, finalLeaf]
        [Multiple \cite{wolf24, mattturner24, luo24}, finalLeaf]
      ]
      [Strength, levelTwo
        [Fixed \cite{panickssery24, bhattacharjee24, mattturner24, jorgensen24, luo24, pham24, li2024inference}, finalLeaf]
        [Adaptive \cite{stoher24, wang24, scalena24, kong24, wang24steering}, finalLeaf]
      ]
      [Intervention Stage, levelTwo
        [All activation spaces \cite{li2024inference, zhu2024language, cao24steering, jorgensen24, pham24, wang24, scalena24, wang24steering, stickland24, wang2024inferaligner, bronzini2024unveiling, heimersheim24}, finalLeaf]
        [Attention activations \cite{bhattacharjee24, stoher24, wang24, xiao24, zhang2024truthx, hu2021lora}, finalLeaf]
        [Residual stream activations \cite{subramani22, zou2023representation, panickssery24, mattturner24, singh24, hendel2023context, li2023transformers, yang2025task, zhuang2024vector, ilharco2022editing, bowen2024beyond, pham2024robust, panigrahi2023task, devin2019plan, huang2024multimodal, luo2024task, stoher24, kong24, dong24, chalnev24, ferrando24halluc, postmus24}, finalLeaf]
        [Target MLP layer activations \cite{subramani22, zou2023representation, luo24, cao2024nothing, stoher24, kong24, wu24, liu2024aligning, zhao24, rahn24, zou24circuit}, finalLeaf]
        [Target neurons \cite{tlaie24, xiao24, wang24, kong24, zhang2024truthx}, finalLeaf]
        [Modifies model(s) \cite{chen24, garg24, choi24, hu2021lora, zhang24, feng24}, finalLeaf]
      ]
    ]
  ]
\end{forest}
\caption{Graph showing Representation Control, with goals that are implemented using specific methods.}
\label{fig:representation_control}
\end{figure}
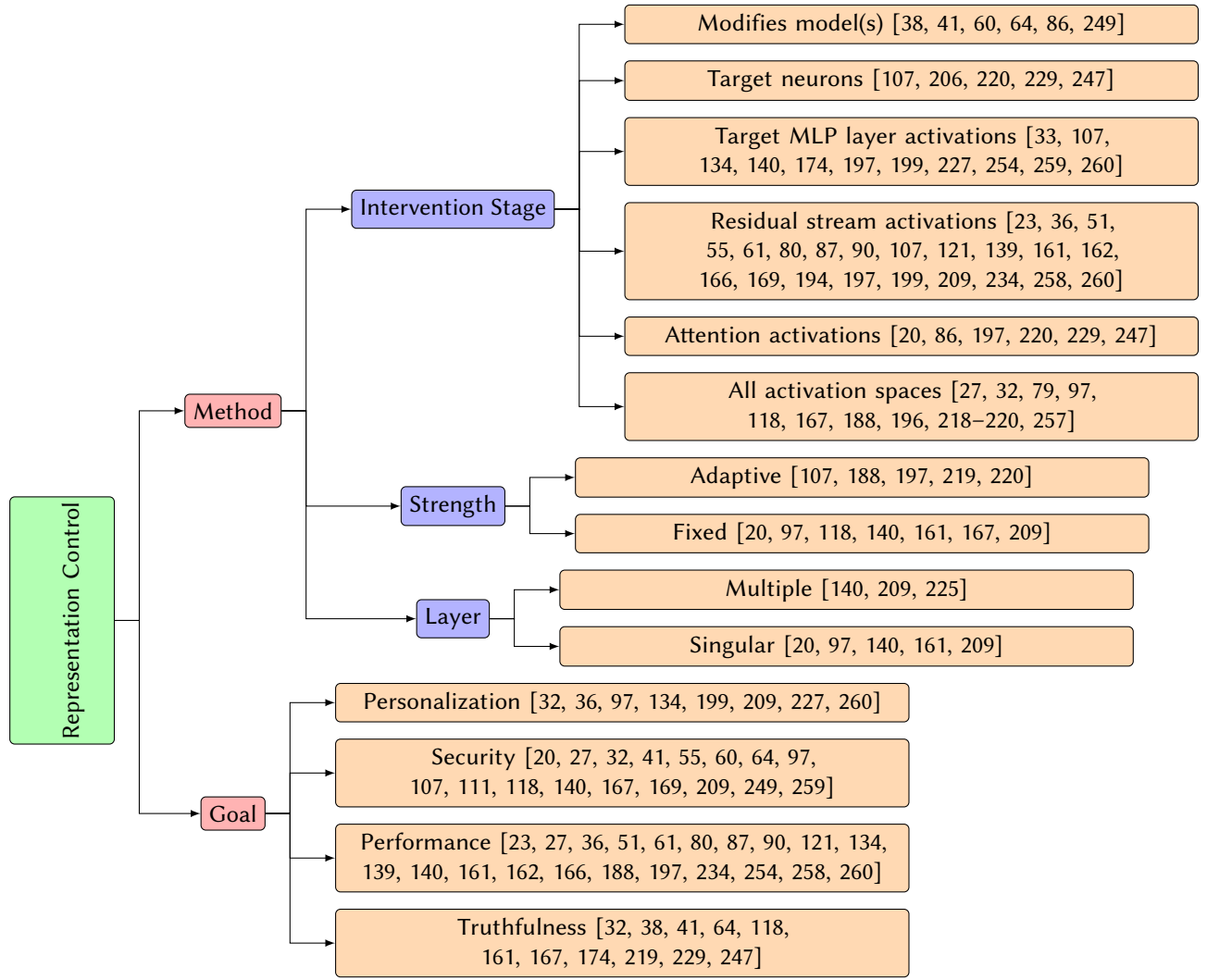

\subsection{Intervention Methods}

\subsubsection{Single and Multi Property}

The representations identified by representation reading techniques can be applied to steer the model into a particular direction. However, steering the model in multiple directions proves difficult and not all methods support applying multiple directions at the same time. For example, steering the model to be both more trustworthy and less prone to jailbreaks requires support for steering the model in multiple directions at the same time.

Theoretically, a representation can be reframed as a more general one, but identifying such representation is more difficult and leads to unpredictible behaviour, usually leading to loss of fluency and generation of nonsensical tokens. Experiments show combining more general representations are usually less effective than combining several injections at different layers \cite{weij24}. Because of this, we define a separation between activation steering techniques that allow for steering the model in a single direction \cite{mattturner24} versus multiple directions \cite{scalena24, cao24steering, weij24, stolfo24}.  

Combining steering vectors for different concepts can lead to unpredictable interactions, where the effect of one intervention interferes with another \cite{xiao24, wolf24}. Moreover, simultaneous injection of individual steering vectors at different places in the model appears to be more effective than combined steering \cite{weij24}. Activation steering can simultaneously enforce multiple instruction constraints, such as requiring an LLM to generate JSON-formatted responses while also adhering to word exclusion rules, by applying composite activation vectors during inference \cite{stolfo24}. Alternatively, Sparse Activation Control can identify distinct attention heads and hence separate interventions over multiple properties \cite{xiao24} or conceptor matrices can be used to eliminate the negative interactions between distinct properties interacting with each other \cite{postmus24}. The implementation of representation control can involve modifying a single layer~\cite{jiang2025probe, panickssery24, mattturner24} or multiple layers~\cite{beaglehole25, li2024inference}.

\subsubsection{Constant or Dynamic}

Once a representation has been identified and the vector (or multiple vectors) has been created, it needs to be applied to the model to modify its activation space. In the process, either the representation engineering developer needs to assign intervention strength, choosing a \textbf{Constant} \cite{mattturner24, zou2023representation, panickssery24}; or it is set as \textbf{Dynamic}, i.e. determined by the model at inference time \cite{tlaie24, wang24steering, scalena24, stoher24}. There are several approaches to setting those weights dynamically, either based on a probe and unsupervised clustering \cite{wang24steering}, Kullback-Leibler (KL) divergence \cite{scalena24}, cosine similarity of activation matrices between unsteered and steered models \cite{scalena24} or gradient-based optimization \cite{stoher24}.

\subsubsection{Intervention Stage}

Representation engineering phases can be categorized based on where the modifications are applied within the model. These can be performed on all activation spaces and components \cite{li2024inference, zhu2024language, cao24steering, jorgensen24, pham24, lee24, bronzini2024unveiling, heimersheim24, wang24, scalena24, wang24steering, stickland24, wang2024inferaligner}. Alternatively, the intervention may target the residual stream activations, that is, the hidden states that pass between layers through residual connections \cite{subramani22, zou2023representation, panickssery24, mattturner24, singh24, hendel2023context, li2023transformers, yang2025task, zhuang2024vector, ilharco2022editing, bowen2024beyond, pham2024robust, panigrahi2023task, devin2019plan, huang2024multimodal, luo2024task, chalnev24, ferrando24halluc, postmus24}. Another approach involves modifications to the target MLP layers \cite{subramani22, zou2023representation, luo24, cao2024nothing, rahn24, wu24, liu2024aligning, zhao24, zou24circuit}, specific neurons \cite{tlaie24, xiao24, zhang2024truthx}, or changes to the model through fine-tuning, re-training, or other forms of weight modification \cite{chen24, kong24, garg24, choi24, hu2021lora, zhang24, feng24}.

\subsection{Intervention Goals}

Representation control methods aim to steer the model towards particular outcomes. It is important to evaluate each method through the lens of its original goal, because identification and intervention on one representation does not mean the intervention is going to be successful on another representation type. The precise nature of representations is difficult to define. A representation can range from broad traits such as honesty to specific facts encoded in activation patterns \cite{herrmann2025standards}. However, extracting such representations is challenging because probing methods often fail to generalize, capturing only sample-specific features rather than the true underlying structures \cite{levinstein2024still}. Moreover, attempts to identify representations may inadvertently conflate them with tasks the model is solving rather than isolating the true concept encodings \cite{kumar2022probing}. This difficulty arises because internal structures of LLMs do not necessarily align with human-meaningful categories, making interpretation reliant on indirect techniques with inherent limitations. Defining the sample and representation scope is therefore critical, and the interventions remain specific to a particular representation. Simple concepts like truthfulness are easier to identify than sophisticated ones like humor or appropriateness \cite{rutte24}.

\paragraph{Personalization}

Customizing the values and response style of an LLM is important for increasing adoption, general satisfaction and capability at completing tasks \cite{zhang2024personalization}. Activation steering is a lightweight method to steer intended behavior without the substantial computational resources of fine-tuning, hence allowing for the creation of individualized LLM experiences \cite{cao24steering}. Identifying and applying representations of particular style is challenging, but possible through activation steering \cite{zou2023representation, cao24steering, jorgensen24, liu2024aligning, wu24, chalnev24, tlaie24}. Embedding certain moral values into the model is also possible. 

\paragraph{Security}

LLMs are aligned to reject harmful requests, but adversaries may nonetheless manipulate them into producing harmful outputs, for example through jailbreaks \cite{bertino2021ai}. LLMs are capable of being manipulated into assisting in unethical requests, leaking personal data or giving toxic responses \cite{yi2024jailbreak, liu2024survey, singh24, mattturner24, bhattacharjee24, luo24, pham24, lee24, stickland24, wang2024inferaligner, dong24, garg24, choi24, kong24, scalena24, zou24circuit, zhang24, feng24}. In particular, circuit breakers have proven to be particularly effective, relying on identification of harmful outputs and preventing them from generating a response \cite{zou24circuit}. Representation engineering can be used to create defenses, but also circumvent existing safety measures for red-teaming purposes. Activation steering can extract unlearned information from large-language models (LLMs), demonstrating vulnerabilities in current unlearning techniques \cite{seyitoglu24}. Malicious attack vectors can be inserted at inference time to steer the model towards harmful behaviour and manipulate model alignment on the representation level, as described in the Trojan Activation Attack \cite{wang24trojan}, future backdoor trigger method \cite{price24},  the Safety Concept Activation Vector (SCAV) \cite{xu24} or direct interventions in the model's representation space to reduce refusal rates \cite{zou24circuit}. Similarly, Li et al. \cite{li24} detect safety patterns in the latent space, which when mitigated indicate that models are less able to refuse harmful requests and strengthening safety patterns indicate that models are more resistant to jailbreaking attempts. One can also create representation-level roleplay jailbreaks \cite{ghandehariom24}. 

Security risks exist not only from the adversaries, but also the model itself. As LLMs become more capable, it becomes difficult to know when a model is pursuing goals that are aligned with its human evaluators, or engaging in strategic deception and faking alignment \cite{greenblatt2024alignmentfakinglargelanguage}. Early experiments suggest that linear probes alone are not sufficient to detect such responses.\cite{goldowskydill2025detectingstrategicdeceptionusing}. 

\paragraph{Performance}

Modern AI training is optimized for enabling the most effective learning algorithm to generate accurate responses to variety of general questions. Interventions in the latent space can steer the model towards better responses. Representations of particular skills, like Chain-of-Thought thinking \cite{zhang24latent}, code type prediction \cite{lucchetti24} or knowledge selection accuracy \cite{zhao24}. Several approaches have been applied to selectively steer the model towards more performant responses at particular tasks \cite{zou2023representation, cao2024nothing, hendel2023context, bronzini2024unveiling, rahn24, stoher24, scalena24, wu24, zhao24, postmus24, liu2024aligning}. In particular, task vectors, a small scale intervention stemming from In-Context-Learning has been successfully applied to steering the model to be better at narrow tasks. 

\paragraph{Truthfulness}

LLM hallucination refers to models generating false outputs with high confidence \cite{salvagno2023artificial}. They are a problem because they can lead to malfunctioning in critical use cases, misinformation and loss of trust in AI \cite{huang2023survey, zhang2023siren}. Improving truthfulness through representation engineering is a major area of research  \cite{pham24,panickssery24,wang24steering,wang24steering,cao24steering, li2024inference, stickland24}. By identifying and amplifing the "truth" direction, activation steering methods have shown to reduce hallucination rates. Examples include HPR \cite{pham24}, CAA \cite{panickssery24} or ACT \cite{wang24steering}. This is primarily achieved by identifying representations from a part of TruthfulQA, a commonly used hallucination benchmark and testing on a held-out set of this benchmark \cite{pham24} \cite{panickssery24}. 

\section{Taxonomy}

\begin{table}[htbp]
    \small
    \centering
    \begin{tabularx}{\textwidth}{>{\raggedright\arraybackslash}X>{\raggedright\arraybackslash}p{2.6cm}>{\raggedright\arraybackslash}X>{\centering\arraybackslash}p{0.1cm}}
        \toprule
        \textbf{Technique Name} & \textbf{Intervention Stage} & \textbf{Goal} & \textbf{Performance} \\
        \midrule
        \rowcolor{linearcontrastive!50} Latent Steering Vectors \cite{subramani22} & [R][T] & Personalization & \harveyBallHalf \\
        \rowcolor{linearcontrastive!50} RepE \cite{zou2023representation} & [R] [T] & Personalization, Performance, Truthfulness  & \harveyBallHalf \\
        \rowcolor{linearcontrastive!50} CAA \cite{panickssery24} & [R] & Truthfulness  & \harveyBallHalf \\
         \rowcolor{linearcontrastive!50} ITI \cite{li2024inference, zhu2024language} & [A] & Truthfulness  & \harveyBallHalf \\
        \rowcolor{linearcontrastive!50} ActAdd \cite{mattturner24} & [R] & Security, Personalization  & \harveyBallHalf \\
        \rowcolor{linearcontrastive!50} C-WSV \cite{bhattacharjee24} & [AA] & Security  & \harveyBallHalf \\
        \rowcolor{linearcontrastive!50} Bi-directional Pref. Optim. \cite{cao24steering} & [A] & Personalization & \harveyBallFull \\
        \rowcolor{linearcontrastive!50} PaCE \cite{luo24} & [T] & Security  & \harveyBallFull \\
        \rowcolor{linearcontrastive!50} Mean-Centered Activ. Steer. \cite{jorgensen24} & [A] & Personalization, Security  & \harveyBallThreeQuarter \\
        \rowcolor{linearcontrastive!50} Household. Pseudo-Rotation \cite{pham24} & [A] & Security, Truthfulness  & \harveyBallFull \\
        \rowcolor{linearcontrastive!50} MiMiC \cite{singh24} & [R] & Security  & \harveyBallHalf \\
\rowcolor{linearcontrastive!50} SARA \cite{tlaie24} & [N] & Personalization  & \harveyBallHalf \\

        \rowcolor{linearcontrastive!50} CAST \cite{lee24} & [A] & Security  & \harveyBallHalf \\
        \rowcolor{linearcontrastive!50} SCANS \cite{cao2024nothing} & [T] & Performance  & \harveyBallHalf \\
        \rowcolor{linearcontrastive!50} Task Vectors \cite{hendel2023context, li2023transformers, yang2025task, zhuang2024vector, ilharco2022editing, bowen2024beyond, pham2024robust, panigrahi2023task, devin2019plan, huang2024multimodal, luo2024task} & [R] & Performance  & \harveyBallHalf \\
        \rowcolor{linearcontrastive!50} Activation Patching \cite{bronzini2024unveiling, heimersheim24} & [A] & Security, Performance, Truthfulness  & \harveyBallFull \\
        \rowcolor{linearcontrastive!50} GRATH \cite{chen24} & [M] &  Truthfulness  & \harveyBallFull \\
        \rowcolor{linearcontrastive!50} Entropic Activation Steering \cite{rahn24} & [T] & Performance  & \harveyBallHalf \\
        \rowcolor{linearcontrastive!50} SAC \cite{xiao24} & [AA][N] &  Truthfulness  & \harveyBallFull \\
        \rowcolor{dynamicstrength!50} Activation Scaling 
        \cite{stoher24} & [AA][R][T] &  Performance  & \harveyBallThreeQuarter \\
        \rowcolor{dynamicstrength!50} SADI \cite{wang24} & [A][AA][N] & Performance, Truthfulness  & \harveyBallThreeQuarter \\
        \rowcolor{dynamicstrength!50} DAC \cite{scalena24} & [A] & Security, Performance  & \harveyBallFull \\
        
        \rowcolor{dynamicstrength!50} RE-CONTROL \cite{kong24} &  [R][N][T] & Security & \harveyBallThreeQuarter \\

        \rowcolor{dynamicstrength!50} ACT \cite{wang24steering} & [A] &  Truthfulness  & \harveyBallHalf\\
         \rowcolor{multiplemodel!50} KL-Then-Steer (KTS) \cite{stickland24} & [A] & Security, Performance  & \harveyBallHalf \\
         \rowcolor{multiplemodel!50} InferAligner \cite{wang2024inferaligner} & [A] & Security  & \harveyBallHalf \\
         \rowcolor{multiplemodel!50} CONTTRANS \cite{dong24} & [R] & Security  & \harveyBallHalf \\
         \rowcolor{multiplemodel!50} KTCR \cite{garg24} & [M] & Security  & \harveyBallHalf \\
        \rowcolor{finetuningalternatives!50} TruthX \cite{zhang2024truthx} & [AA][N]  &  Truthfulness  & \harveyBallHalf \\
        \rowcolor{finetuningalternatives!50} Safety-Aware Fine-tuning \cite{choi24} & [M] & Security  & \harveyBallHalf \\
        \rowcolor{finetuningalternatives!50} ReFT \cite{wu24} & [T] & Personalization, Performance  & \harveyBallFull \\
        \rowcolor{finetuningalternatives!50} RAHF \cite{liu2024aligning} & [T] & Personalization, Performance, Truthfulness  & \harveyBallFull \\
        \rowcolor{finetuningalternatives!50} LoRRA \cite{hu2021lora} & [M][AA] & Personalization, Performance, Truthfulness  & \harveyBallHalf \\ 
         \rowcolor{sparseautoencoders!50} SPARE \cite{zhao24} & [T] & Performance & \harveyBallHalf \\
        
         \rowcolor{sparseautoencoders!50} SAE-TS \cite{chalnev24} & [R] & Personalization & \harveyBallHalf \\
         \rowcolor{sparseautoencoders!50} Self-knowledge SA \cite{ferrando24halluc} & [R] & Truthfulness  & \harveyBallHalf \\

        ARE \cite{zhang24} & [M] & Security, Truthfulness  & \harveyBallFull \\
        Conceptors \cite{postmus24} & [R] &  Performance  & \harveyBallFull \\
        Circuit Breakers \cite{zou24circuit} & [T] & Security  & \harveyBallFull \\
 
        LEGEND \cite{feng24} & [M] &  Security & \harveyBallQuarter \\
        \bottomrule
    \end{tabularx}
    
    \vspace{0.5cm}
\begin{center}
\begin{tikzpicture}[font=\scriptsize]
  \node[draw, fill=linearcontrastive, minimum size=0.35cm] (box1) at (0, 0) {};
  \node[right=0.1cm of box1] {Linear Fixed Methods};
  
  \node[draw, fill=multiplemodel, minimum size=0.35cm] (box2) at (0, -0.6) {};
  \node[right=0.1cm of box2] {Multiple Model Methods};
  
  \node[draw, fill=dynamicstrength, minimum size=0.35cm] (box3) at (0, -1.2) {};
  \node[right=0.1cm of box3] {Dynamic Strength Methods};
  
  \node[draw, fill=finetuningalternatives, minimum size=0.35cm] (box4) at (4, 0) {};
  \node[right=0.1cm of box4] {Fine-tuning Methods};
  
  \node[draw, fill=sparseautoencoders, minimum size=0.35cm] (box5) at (4, -0.6) {};
  \node[right=0.1cm of box5] {Sparse Auto-Encoders};
  
  \node[draw, fill=white, minimum size=0.35cm] (box6) at (4, -1.2) {};
  \node[right=0.1cm of box6] {Other};
  
  \node[anchor=west] at (8, 0) {[A]: All activation spaces};
  \node[anchor=west] at (8, -0.6) {[AA]: Attention activations};
  \node[anchor=west] at (8, -1.2) {[R]: Residual stream activations};
  
  \node[anchor=west] at (11.5, 0) {[T]: Target MLP layer activations};
  \node[anchor=west] at (11.5, -0.6) {[N]: Target neurons (+/- activations)};
  \node[anchor=west, align=left, text width=4cm] at (11.5, -1.2) {[M]: Modifies model(s)\\(fine-tuning, re-training, etc.)};
\end{tikzpicture}
\end{center}
    
    \caption{Taxonomy of Representation Engineering Methods.}
    \label{tab:steering-techniques}
\end{table}

We structure all identified representation engineering methods in a comprehensive taxonomy. We provide a relative ranking of these methods based on their performance. For performance, we evaluate whether these methods have been proven by subsequent research to be less effective, whether issues lead to fluency degradation, and how comprehensive the original testing across multiple representations has been.

\subsection{Linear Contrastive Steering Vector}

Linear contrastive steering vectors provide a framework for modifying transformer model behavior through targeted interventions in the hidden activation space. These methods leverage differences in activation patterns—obtained by contrasting selected prompt pairs or decomposing internal representations—to derive vectors that steer model outputs toward desired behaviors.

\subsubsection{Simple Contrastive Vectors}

The contrastive steering vector is computed as the difference between representations from a target and reference scenario:
\begin{equation}
v_c = \text{Rep}(M, T^+ ) - \text{Rep}(M, T^- )
\end{equation}
where $T^+$ and $T^-$ correspond to positive and negative stimuli for the concept under study and $M$ is the model. The intervention is applied by modifying the original representation through a controlled addition or subtraction of the contrastive vector:
\begin{equation}
R' = R + \alpha v_c
\end{equation}
where $\alpha$ is a scaling coefficient that adjusts the strength of the intervention. This common framework has been developed in parallel, spanning common methods across RepE \cite{zou2023representation}, CAA \cite{panickssery24}, ITI \cite{li2024inference} and ActAdd \cite{mattturner24}. 

\paragraph{CAA (Mean Difference)}  

Contrastive Activation Addition (CAA), modifies activations in the residual stream \cite{panickssery24}. CAA constructs steering vectors by averaging the difference in activations between contrastive prompt pairs, such as factual versus hallucinatory responses, at a specific transformer layer. These vectors are then applied at inference time to modulate model outputs, either reinforcing or suppressing target behaviors like sycophancy, corrigibility, or refusal. CAA outperforms system-prompting and shows complementary effects with finetuning. CAA vectors exhibit layer-wise consistency, transferability between base and fine-tuned models, and alignment with behaviorally meaningful activation clusters, improving their interpretability. 

\begin{equation}
v_c = \frac{1}{N} \sum_{i=1}^{N} \left( \text{Rep}(M, T^+_i) - \text{Rep}(M, T^-_i) \right)
\end{equation}

\paragraph{RepE (PCA-Based)}  

Representation Engineering (RepE) transforms internal representations using vectors derived from Linear Artificial Tomography (LAT) \cite{zou2023representation}. It identifies directions in representation space that correlate with cognitive functions, then intervenes by linearly combining these reading vectors with the model's activations. This is achieved through two methods: (1) Reading Vector Intervention, where a fixed direction is added or subtracted from activations to amplify or suppress the associated cognitive function, and (2) Contrast Vector Intervention, where contrast vectors are calculated by running paired contrastive prompts through the model and subtracting their activations. 

\begin{equation}
v_c = \text{TopPC} \left( \{ \text{Rep}(M, T^+_i) - \text{Rep}(M, T^-_i) \}_{i=1}^{N} \right)
\end{equation}

\paragraph{ITI (Classifier-Based)}  

Inference-Time Intervention (ITI) enhances the truthfulness of language models by shifting activations in specific directions during inference \cite{li2024inference, zhu2024language}. It identifies attention heads with high linear probing accuracy for truthfulness and intervenes only in these heads by shifting activations along directions correlated with truthful outputs. This intervention is applied autoregressively throughout the text generation process. ITI utilizes two main parameters: the number of attention heads to adjust and the magnitude of the shift, enabling fine-grained control over the intervention. Unlike weight editing methods, ITI operates directly on attention head activations, preserving the underlying model weights and reducing computational costs.

\begin{equation}
v_c = w^*, \quad A'_h = A_h + \alpha v_c
\end{equation}

\paragraph{ActAdd (Single Pair)}  

Activation Addition (ActAdd) computes a steering vector by contrasting activations from prompt pairs, such as "Love" vs. "Hate," and applies this vector during inference to modify model behavior \cite{mattturner24}. Most importantly, it can achieve good results with even one contrastive pair, and the effects use a simple subtraction of contrasting activations to create the steering vector. 

\begin{equation}
v_c = \text{Rep}(M, T^+) - \text{Rep}(M, T^-)
\end{equation}

Studies examining these have found CAA to be most reliable and also formulated theory as to why this is the optimal method \cite{im2025unified}.

\paragraph{LSV}
Latent Steering Vectors (LSVs) intervene in pretrained language models by adding fixed-length vectors, $ z_{\text{steer}} $, directly to the hidden states during decoding, influencing model outputs without altering model weights \cite{subramani22}. Specifically, $ z_{\text{steer}} $ is optimized via gradient descent to maximize the log probability of a target sentence, keeping the language model frozen. 

\[
z_{\text{steer}} = \arg\max_{z} \sum_{t=1}^{T} \log p(x_t | x_{<t}, z_{\text{steer}})
\]

LSVs can be injected at different layers, including embedding, self-attention, feed-forward, and the final language model head, but are most effective when added to the middle layers (e.g., layers 6 or 7 in GPT-2). These vectors are injected either at every timestep or just the first timestep, with minimal performance loss in the latter case. By linearly combining $ z_{\text{steer}} $ with the model’s hidden states, LSVs steer the model to generate desired outputs with high accuracy. This method preserves the model's overall behavior since it operates at the activation level rather than modifying weights or retraining.

\[
h' = h + \alpha \cdot z_{\text{steer}}
\]

Latent steering vectors can be extracted directly from pretrained language models without requiring fine-tuning \cite{subramani22}. These vectors encode control mechanisms within the model’s activations, allowing direct manipulation of text generation. By injecting steering vectors into hidden states, models can achieve near-perfect target sentence recovery and enable unsupervised sentiment transfer on par with specialized models. Notably, steering vectors preserve sentence similarity relationships and outperform traditional hidden-state pooling methods in semantic textual similarity benchmarks. Their effectiveness depends on the injection location within the transformer stack, with middle layers providing optimal results—consistent with findings in adapter-based fine-tuning. Importantly, these vectors do not merely memorize sequences but encode generalizable latent structures, reinforcing their potential for controlled text generation and precise representation manipulation.

\paragraph{C-W SV}
Category-wise Steering Vectors \cite{bhattacharjee24} compute steering vectors by computing activation differences of harmful and harmless text and applying these vectors at specific model layers to redirect outputs toward safer regions in latent space. This approach includes both unsupervised and guided methods for vector extraction, with the latter incorporating external safety classifiers to refine steering signals.

\subsubsection{Optimized Steering Vectors}

Recent literature aims to optimize the simple steering vectors, increasing their effectiveness for particular use cases. This section presents the methods with the highest empirical validation to show alternatives to simple contrastive vector interventions. 

\paragraph{BiPO}
Bi-directional Preference Optimization (BiPO) optimizes steering vectors by adjusting the generation probability of human preference data pairs instead of relying on direct activation differences from contrastive prompts \cite{cao24steering}. BiPO has shown to be particularly effective at steering AI personas compared to other representation engineering techniques. BiPO scored higher than Contrastive Activation Addition and Freeform on every personalization task it was evaluated on. This approach also improves the model performance on tasks such as truthfulness, hallucination mitigation, and jailbreaking defense. The vectors calculated using this approach transfer across models and fine-tuned LoRAs. BiPO allows for the application of multiple steering vectors to influence multiple behaviors simultaneously without decreasing the general capabilities of the model.

\paragraph{PaCE}

Parsimonious Concept Engineering (PaCE) constructs a large-scale concept dictionary in the activation space and uses sparse coding techniques to decompose model activations into benign and undesirable components \cite{luo24}. By using oblique projection and a sparse linear combination of concept directions, a new vector is constructed embedding all beneficial concepts in it. PaCE is shown to maintain fluency while preserving safe intervention effect. 

\paragraph{Mean-centering}
Mean-centering improves activation steering by refining steering vectors through a subtraction operation: the mean activation of a training dataset is subtracted from the mean activation of a target dataset \cite{jorgensen24}. This removes global biases in activations and improves control over model outputs. Applying mean-centering to toxicity reduction in language models results in lower toxicity scores compared to existing activation steering methods. Mean-centering increases the relevance of genre-specific terms when steering a model to continue a story in a target genre. It also improves the accuracy of function vectors, such as those used to extract country-capital relationships \cite{jorgensen24}. 

\paragraph{HPR}
Householder Pseudo-Rotation (HPR) treats activations as having both direction and magnitude, ensuring norm preservation during edits \cite{pham24}. This is fundamentally different from interventions that simply perform addition \cite{panickssery24}. HPR first learns a hyperplane separating desirable and undesirable activations, then reflects negative activations across this plane using a Householder transformation, followed by a rotation to align them with positive behaviors. Experiments show that HPR significantly outperforms other methods in hallucination reduction. This approach significantly improves over the previous linear additive steering vectors, also by proving the evidence of the Magnitude Consistency Property. 

The magnitude consistency property means that the magnitude of a representation vector remains relatively stable when the underlying semantic meaning of the input does not change fundamentally \cite{pham24}. This property helps to \emph{a)} detect meaningful semantic representations by observing how vector magnitudes behave under different transformations
, \emph{b)} identify when neural networks are developing robust, consistent internal representations, and \emph{c)} understand potential failure modes where representations might become unstable or distorted. For example, if you input two semantically similar sentences into a language model, a magnitude-consistent representation would show similar vector magnitudes, indicating the model has captured a stable semantic understanding rather than arbitrary mappings. This property highlights a key limitation of the point-in-space view, where the steering vector approach cannot simultaneously maintain activation magnitude consistency and effectively edit the activation to achieve greater performance improvement for desired behaviors for language models \cite{pham24, park24}.

\begin{figure}[h]
  \centering
  \includegraphics[width=\linewidth]{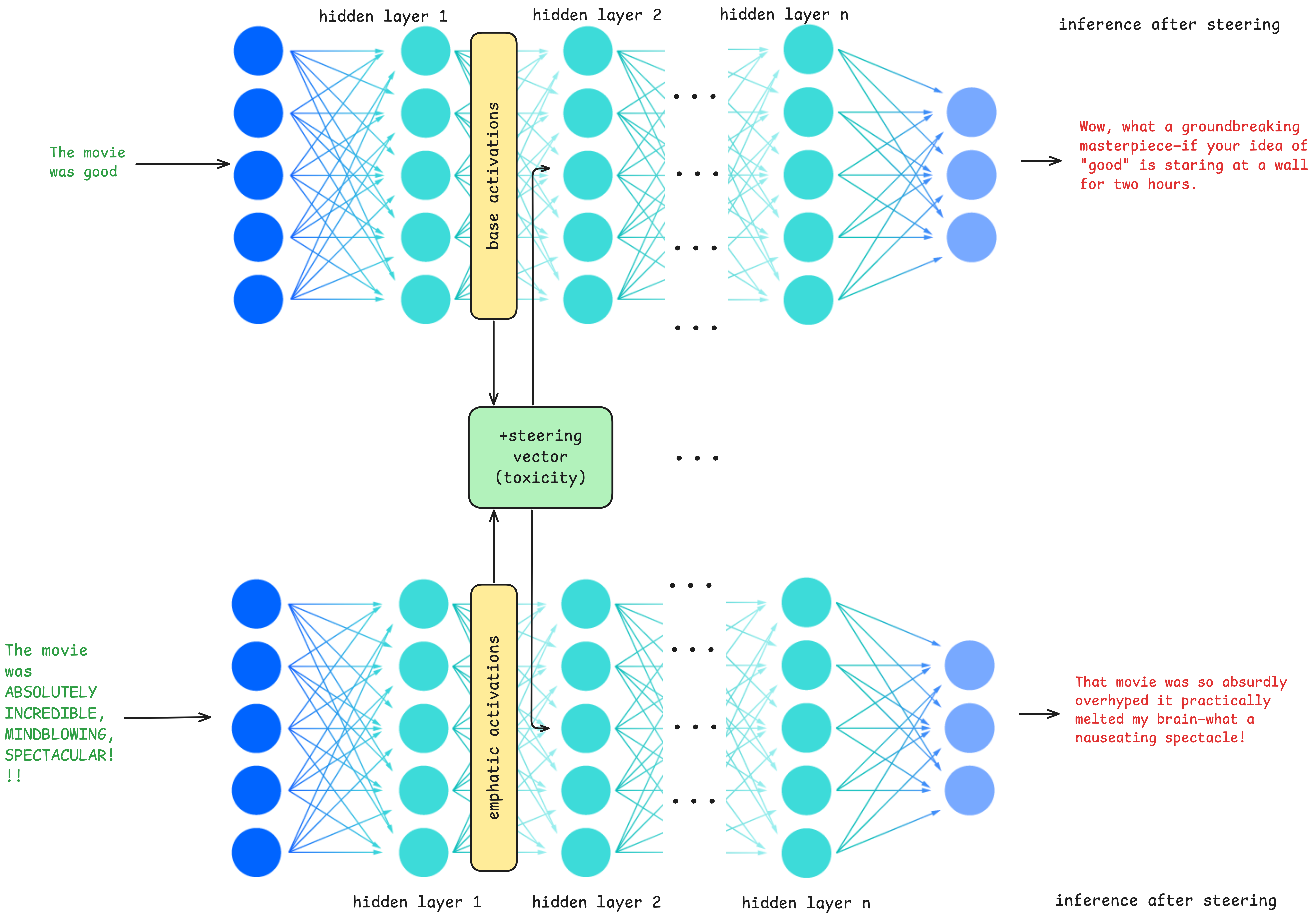}
  \caption{Violation of the magnitude consistency property (high-level view). In order to uphold the magnitude consistency property, a robust LLM should have the same (or nearly equal) activation vector norms for semantically similar inputs. Steering vector intervention disrupts that by causing disproportionate scaling of norms and context collapse. In this example, the activation vector norm of the base input and the emphatic input will not remain the same after a steering vector that encompasses toxicity is applied.}
\end{figure}

\paragraph{MiMiC}

Minimally Modified Counterfactuals (MiMiC) is an affine steering method that shifts language model representations toward a target concept by modifying their mean and covariance \cite{singh24}.
MiMiC achieves this by applying a linear transformation that aligns the concept-conditional distributions, ensuring that representations from different groups become statistically similar.
Unlike simple mean-matching approaches, MiMiC also mitigates bias by neighbors, meaning that the spatial clustering of representations no longer correlates with protected attributes like gender or dialect.

\paragraph{SARA}

Similarity-based Activation Steering with Repulsion and Attraction (SARA) is a causal intervention technique that modifies the internal activations of a language model to align with specific ethical perspectives while suppressing undesired ones \cite{tlaie24}. It operates by computing cosine similarity between activation matrices of different prompts, selectively reinforcing target activations and repelling conflicting ones through singular value decomposition (SVD). Unlike standard activation steering, which applies uniform shifts, SARA dynamically adjusts activations based on their initial similarity to the target, ensuring more precise and context-aware modifications.

\paragraph{CAST}
Conditional Activation Steering (CAST) introduces condition vectors that selectively determine when steering happens, allowing models to refuse specific types of inputs without affecting unrelated responses \cite{lee24}. CAST relies on cosine similarity between a model’s hidden state and predefined condition vectors to trigger interventions. This means that refusal behavior is applied only in relevant contexts. Experiments show it results in better security and performance than simple activation steering. The method also enables logical composition of condition vectors, such as enforcing refusals for legal or medical inquiries while permitting general responses. 

\paragraph{Probe-Free Low-Rank Activation Intervention (FLORAIN)}
FLORAIN \cite{jiang2025probe} introduces a probe-free, low-rank activation intervention to modify language model activations at inference time, improving the generation of desirable content. FLORAIN directly learns a low-rank transformation that projects activations onto a desirable manifold modeled as an ellipsoid region. This approach ensures efficient inference-time interventions, reduces computational overhead by operating on a single layer, and minimizes representation shifts across layers. To address the challenge of high-dimensional activation vectors with limited training samples, FLORAIN employs an extrapolation strategy that shifts mean activation vector away from undesirable activations by combining a question-specific adjustment (capturing the difference between desirable and undesirable activations for a given question) and a global adjustment (capturing the overall shift from undesirable to desirable activations across all data). A weighting factor balances the influence of local versus global information, ensuring robustness even in low-data scenarios. This refined mean activation vector defines the ellipsoid-shaped desirable activation region onto which activations are projected during inference, enhancing truthfulness and coherence in generated outputs.

\paragraph{Aggregate and Conquer: Detecting and Steering LLM Concepts}
This method \cite{beaglehole25} detects and steers LLM concepts by leveraging Recursive Feature Machines (RFMs) to extract eigenvector-based concept representations from activations. For detection, it trains layer-wise predictors to classify activations, then computes the Average Gradient Outer Product (AGOP) matrix to identify the most influential directions, selecting its top eigenvectors as concept vectors and aggregating them across layers for improved accuracy. For steering, it modifies activations during the forward pass by replacing each layer’s activation with the learned concept vector and a control coefficient that varies per concept and model. To steer multiple concepts simultaneously, it combines layer-wise concept vectors in a weighted sum, allowing fine-grained control over behaviors such as reducing toxicity, enhancing truthfulness, altering writing styles, or modifying sentiment strength. Unlike previous methods, this approach learns concept vectors from nonlinear predictors, enables multi-layer and multi-concept steering, and supports gradated control, making it a powerful and efficient alternative to fine-tuning or contrastive interventions.

\subsubsection{In-Context Learning and Task Vectors}

In-context learning (ICL) enables large-language models to adapt dynamically to new tasks by leveraging additional examples to modify internal representations without parameter updates by providing few-shot examples of answers to the task \cite{dong2022survey, brown2020language, li2023transformers}. Attention mechanisms in transformers can be viewed as performing a form of implicit gradient descent, effectively allowing the model to optimize for new tasks on the fly \cite{dai2022can, von2023transformers}. This can be formalized using PAC learning frameworks to establish finite sample complexity bounds \cite{wies2023learnability}. ICL can be viewed as an implicit learning algorithm, where transformers encode smaller models within their activations and adjust them as new examples are introduced \cite{akyurek2022learning, park2024iclr}.  ICL can also improve model performance through self-verification \cite{wang2023large}. The effectiveness of ICL can be explained by implicit Bayesian inference, where the model infers a latent structure that connects the pretraining and inference distributions \cite{xie2021explanation}.

ICL utilises specialized attention heads (induction heads) that detect and replicate patterns in token sequences, effectively driving the model’s ability to generalize from context \cite{olsson2022context}.  A subset of induction heads, semantic induction heads encode structured relationships such as syntactic dependencies and knowledge graph associations showing how additional examples at inference change representation structure \cite{ren2024identifying}. ICL can successfully approximate complex function classes, including linear models, decision trees, and even neural networks, showing that transformers can implement efficient learning algorithms internally \cite{garg2022can, liu2023icv}. ICL alters embeddings and attention distributions to prioritize task-relevant information while reducing sensitivity to adversarial distractions, also in the case where the provided examples are not particularly useful for task performance \cite{yousefi2023context}. Some studies show that example-label correctness has minimal impact while others find significant sensitivity to correct label assignments, showing that the effectiveness of ICL depends on dataset structure and model scale \cite{yoo2022ground, min2022rethinking}. It is the structure of demonstrations that is important, as models can leverage input distribution patterns and sequence formats even when labels are randomized, showing that task representation might matter more than exact label correctness \cite{min2022rethinking, reynolds2021prompt}. Increasing the number of examples leads to significant performance gains across diverse tasks \cite{agarwal2024many}. While ICL consistently improves task performance over zero-shot prompting, its effectiveness is highly dependent on the selection and ordering of examples, as variations in these factors can cause substantial fluctuations in model accuracy \cite{liu2021makes, zhao2021calibrate}. In-context learning can help the model identify the right parts of the input space and the overall format of the sequence \cite{min2022rethinking}. ICL is similar to RepE in that it alters the representations. However, it does not monitor the full effect providing extra tokens in the input has throughout the layers. 

ICL can be used to crate a  task vectors are compressed versions of contextual examples that condition model behaviour with a latent representation \cite{hendel2023context, li2023transformers}. Task vectors can be thought of as compact, low-level representation engineering intervention. High-dimensional task embeddings align with how humans mentally structure and manipulate knowledge \cite{piantadosi2024concepts, mu2024learning, shao2023compositional}. Task vectors naturally emerge in transformer-based models during in-context learning, where they encode latent task structures within activation spaces, allowing models to generalize efficiently across similar input contexts without explicit weight updates \cite{hendel2023context, yang2025task, zhuang2024vector}. Task vectors encode task-specific transformations in model weight space, enabling multi-task learning and transferability across different domains \cite{ilharco2022editing, yang2025task, bowen2024beyond}. Task vectors allow models to create, subtract, and merge different tasks. Their robustness can be further improved through selective importance weighting and structured sparsification techniques \cite{pham2024robust, bowen2024beyond, panigrahi2023task}. Task vectors have been also been applied in multi-modal and robotic control settings, where they enable compositional plan representations that facilitate generalization to novel tasks and modalities \cite{devin2019plan, huang2024multimodal, luo2024task}.

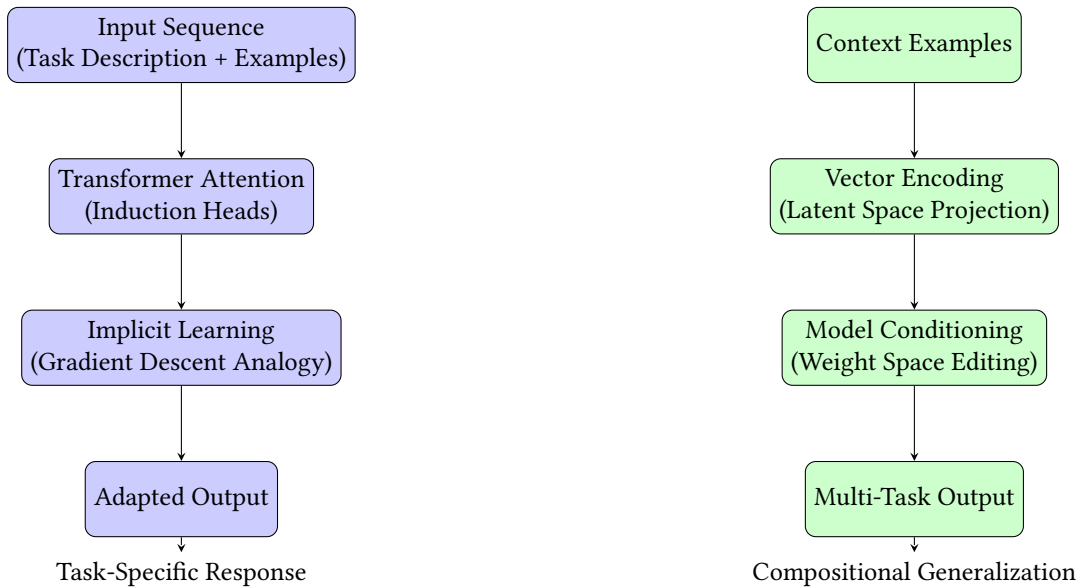
\begin{figure}[t]
\centering
\begin{tikzpicture}[
    node distance=1cm,
    block/.style={draw, rounded corners, minimum width=2.5cm, minimum height=1cm, align=center},
    callout/.style={draw, dashed, rounded corners, fill=white, inner sep=5pt},
    arrow/.style={->, >=stealth},
    comp/.style={fill=blue!20},
    task/.style={fill=green!20}
]

\node[block, comp] (input) {Input Sequence \\ (Task Description + Examples)};
\node[block, comp, below=of input] (attention) {Transformer Attention \\ (Induction Heads)};
\node[block, comp, below=of attention] (implicit) {Implicit Learning \\ (Gradient Descent Analogy)};
\node[block, comp, below=of implicit] (adapted) {Adapted Output};
\node[below=0.2cm of adapted] (icl-out) {Task-Specific Response};

\draw[arrow] (input) -- (attention);
\draw[arrow] (attention) -- (implicit);
\draw[arrow] (implicit) -- (adapted);
\draw[arrow, dashed] (adapted) -- (icl-out);

\node[block, task, right=6cm of input] (context) {Context Examples};
\node[block, task, below=of context] (encoding) {Vector Encoding \\ (Latent Space Projection)};
\node[block, task, below=of encoding] (conditioning) {Model Conditioning \\ (Weight Space Editing)};
\node[block, task, below=of conditioning] (multi) {Multi-Task Output};
\node[below=0.2cm of multi] (task-out) {Compositional Generalization};

\draw[arrow] (context) -- (encoding);
\draw[arrow] (encoding) -- (conditioning);
\draw[arrow] (conditioning) -- (multi);
\draw[arrow, dashed] (multi) -- (task-out);

\end{tikzpicture}
\caption{In-Context Learning (left) modifies model behavior through attention mechanisms and demonstration examples, while task vectors (right) create persistent latent representations that enable direct model conditioning.}
\label{fig:icl_tasks}
\end{figure}

\subsubsection{Activation Patching}

Activation patching (also known as causal tracing) is a method at the intersection of mechanistic interpretability and representation engineering. It seeks to identify small-scale model components responsible for specific behaviors by changing token representations during inference\cite{heimersheim24, bronzini2024unveiling}. Activation patching replaces activations with those from a different inference run. Through this, it traces how specific activations contribute to model behavior. For example, if a model completes \textit{“The Eiffel Tower is in”} with \textit{“Paris,”} activations from an inference with a corrupted prompt (e.g., \textit{“The Colosseum is in” for "Paris"}) to observe the impact on output \cite{zhang2023towards}. The latent representation of this placeholder is replaced with a weighted sum of the input’s latent representations from the original inference, effectively "patching" the activation \cite{yeo2024towards, bronzini2024unveiling}. Noising and denoising are two primary approaches to activation patching: denoising patches clean activations into a corrupted run to identify sufficient components, while noising patches corrupted activations into a clean run to determine necessary components \cite{heimersheim24}. Different corruption methods—such as Gaussian noising and symmetric token replacement—can be implemented to look at changes in interpretability \cite{zhang2023towards}.

\subsubsection{SCANS}

SCANS (Safety-Conscious Activation Steering) extracts "refusal steering vectors" from the model's hidden states by contrasting activations from harmful and benign queries, identifying specific layers responsible for refusal behavior \cite{cao2024nothing}. This is used to prevent the model refusing to follow benign queries from the users. These vectors are then used to adjust activations during inference—steering responses away from unnecessary refusals. SCANS achieves this by first classifying a given query as harmful or benign using a similarity-based method that compares hidden state transitions with a learned reference direction. If a query is deemed benign, SCANS modifies the activations in safety-critical layers to reduce the likelihood of refusal without altering the model’s core capabilities.

\subsubsection{GRATH}

GRATH generates and refines truthfulness training data using out-of-domain questions, then iteratively fine-tunes the model through Direct Preference Optimization (DPO) to improve its alignment with factual correctness \cite{chen24}. What the model essentially does is self-generate pairs of information it then self-truthifies itself on. GRATH significantly improves model accuracy on the TruthfulQA benchmark. GRATH is more cost-effective and scalable than human-annotated datasets for improving the reliability of its outputs. 

\subsubsection{EAST}

AI agents exhibit overconfidence in decision-making, often committing prematurely to actions without sufficient exploration \cite{rahn24}. Entropic Activation Steering (EAST) directly modifies an LLM’s internal uncertainty representation, increasing exploration in sequential decision-making tasks. Unlike adjusting sampling temperature, which minimally affects action entropy, EAST constructs a steering vector by averaging activations weighted by action entropy and applies it at inference time to modulate decision confidence. Experiments in bandit environments demonstrate that EAST significantly increases exploration, prevents premature commitment to suboptimal choices, and generalizes across task variants. EAST modifies both action distributions and model-generated thoughts, shifting them toward expressions of uncertainty and caution, revealing that LLMs encode explicit representations of decision uncertainty that can be directly manipulated.

\subsubsection{SAC}

Sparse Activation Control (SAC) enables multi-dimensional trustworthiness improvements in large-language models (LLMs) by identifying and modifying task-specific attention heads without interfering with general performance \cite{xiao24}. Unlike traditional representation engineering methods that struggle with simultaneous control of multiple behaviors, SAC leverages the sparsity and independence of attention heads to separately regulate safety, factuality, and bias. By employing path patching and Gaussian Mixture Models (GMM), SAC precisely identifies and manipulates the activations of causally relevant components, achieving task-specific control without degrading overall model capabilities. Experimental results on LLaMA models show that SAC outperforms existing methods in multi-task alignment, avoiding the control conflicts that arise when multiple behavioral modifications interact. This work highlights the potential of sparsity-based interventions as a structured and effective alternative to reinforcement learning from human feedback (RLHF) for trustworthiness alignment.
 
\subsection{Dynamic Strength Vectors}

Dynamic strength approaches adjust model activations during inference in a way that changes depending on the input or the stage of generation. Rather than adding a fixed, constant vector of certain magnitude, these methods change the magnitude of hidden activations—sometimes by multiplying them with learned factors—to either magnify or reduce certain signals. These methods adapt the intensity of activation changes to better guide model outputs without compromising overall language ability or adjusting the vectors to deal with a wider variety of interventions. Activation scaling modifies the magnitude of specific activation vectors in language models to steer outputs while preserving interpretability

Stoehr et al. \cite{stoher24} implements activation scaling through methods like SteerVec and ActivScalar  \cite{stoher24}. Unlike additive steering vectors, which alter both direction and magnitude, activation scaling applies learned multiplicative scalars to existing model activations, strengthening or weakening task-relevant representations. The method is optimized using a three-term objective balancing effectiveness (correcting model outputs), faithfulness (minimizing unintended side effects), and minimality (sparsely modifying activations). A dynamic variant, DynScalar, further generalizes activation scaling by learning functions of activation vectors, allowing interventions to adapt to varying prompt lengths without predefined positions. SADI replaces the fixed approach with a dynamic steering vector that adjusts model activations based on contrastive activation differences, allowing targeted interventions at inference time \cite{tlaie24}. It applies binary masking to identify critical model components, such as attention heads and feedforward neurons, and modifies activations based on input semantics without requiring additional training. Dynamic Activation Composition (DAC) enables multi-property steering of LLMs by dynamically adjusting activation steering intensity throughout generation \cite{scalena24}. DAC continuously modulates the strength of multiple steering vectors, balancing conditioning strength with generation fluency. Experiments show that DAC effectively maintains strong conditioning for properties like safety, formality, and language while reducing fluency degradation. RE-CONTROL dynamically edits LLM hidden representations at test time through an approach derived from control theory and the optimal control formulation \cite{kong24}. Instead of modifying model weights through fine-tuning, it perturbs hidden states using learned control signals, which enables efficient alignment with human objectives while maintaining computational efficiency. The method formulates autoregressive language models as discrete-time stochastic dynamical systems and optimizes control signals via a trained value function, allowing for flexible intervention during generation. Empirical results show that RE-CONTROL outperforms existing test-time alignment methods like prompting and guided decoding while generalizing well to new inputs. Adaptive Activation Steering (ACT) dynamically modifies model activations to enhance truthfulness in large-language models (LLMs) without fine-tuning \cite{wang24steering}. ACT adjusts the steering intensity based on the truthfulness content of activations, ensuring targeted intervention. The approach clusters hallucination-related activations and generates distinct steering vectors for different categories of hallucinations. 

\subsection{Specific Representation Engineering Implementation Choices}

Representation engineering relies on the choice of technique and specific hyperparameters. Because of this, there remain several aspects separating the existing approaches from each other. While the theoretical, general approach to RepE implementation is clear, in practice the approaches differ in actual implementation. 

\begin{figure}
  \centering
  \begin{minipage}{0.6\textwidth}
      \centering
      \includegraphics[width=1\textwidth]{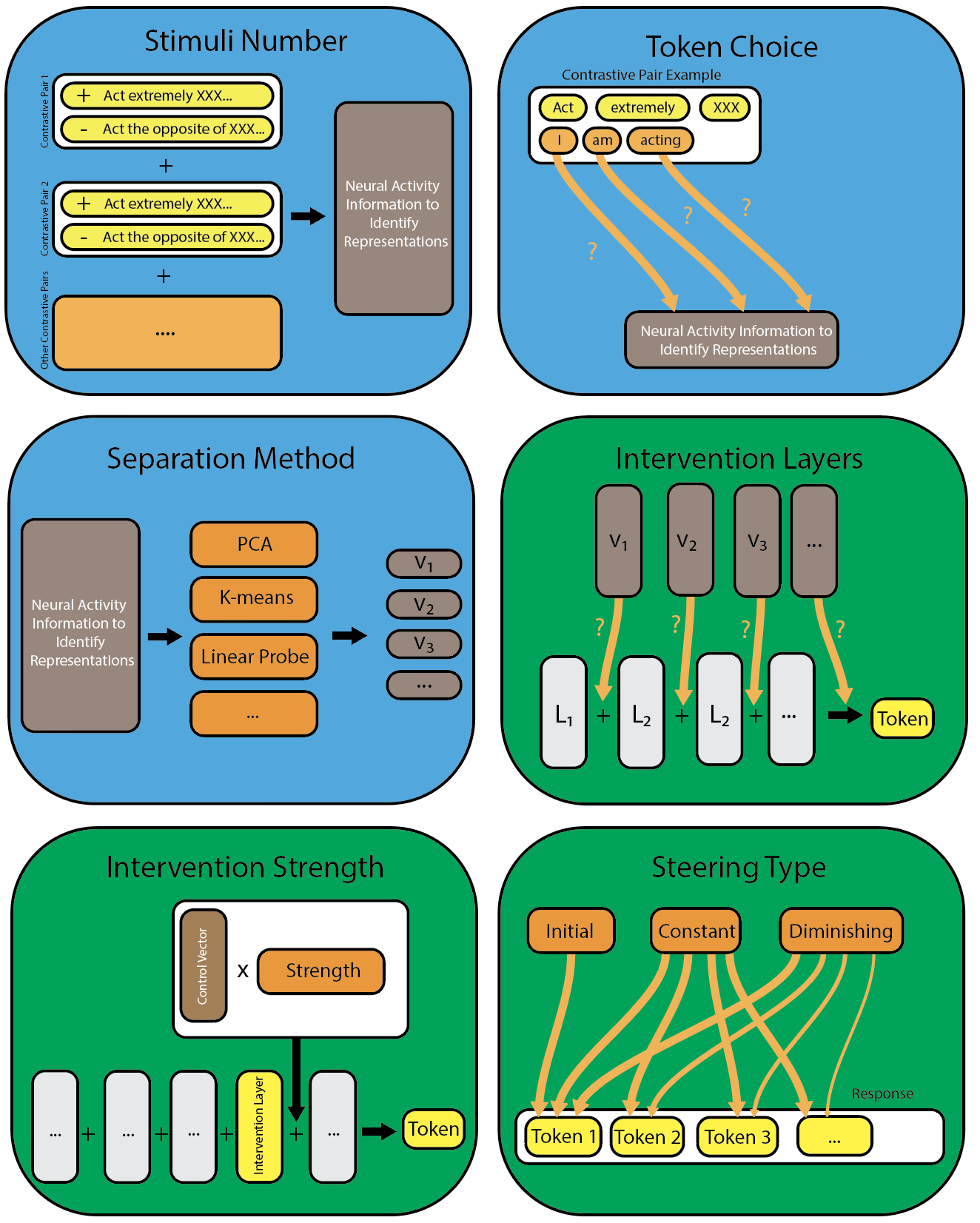}
  \end{minipage}

  \begin{minipage}{0.48\textwidth}
    \vspace{3mm}
    \begin{tikzpicture}
      \node[draw, fill=myorange, minimum size=0.4cm] (box1) at (0,0) {};
      \node[right=0.5cm of box1] {Representation Engineering Parameters};
      
      \node[draw, fill=myblue, minimum size=0.4cm] (box2) at (0,-0.6cm) {};
      \node[right=0.5cm of box2] {Representation Reading Parameters};
      
      \node[draw, fill=mygreen, minimum size=0.4cm] (box3) at (0,-1.2cm) {};
      \node[right=0.5cm of box3] {Representation Control Parameters};
    \end{tikzpicture}
  \end{minipage}
  
  \caption[Representation of Parameters]{%
    Representation of Parameters.}
  \Description{Representation of Parameters}
\end{figure}
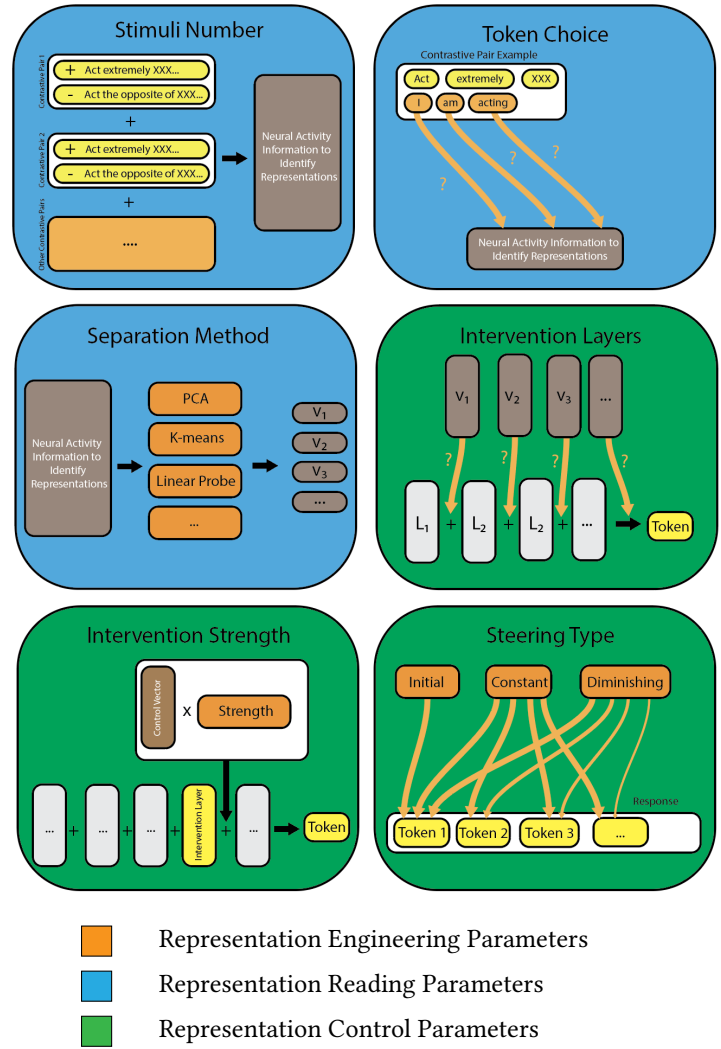

\subsubsection{Stimuli Number}

Many methods use benchmarks to construct the steering vector and utilize the remainder of it to test for effectiveness, operating under the assumption that the larger the number of stimuli the better \cite{panickssery24}. However, in practice, a higher number of stimuli incurs extra processing cost while improving accuracy only slightly \cite{zou2023representation}. The number of stimuli varies widely in literature, ranging from as few as one contrastive pair \cite{mattturner24}, 5-128 strings \cite{zou2023representation}, 30 prompt pairs \cite{scalena24}, 64 questions \cite{cao2024nothing}, 100 prompts \cite{rahn24}, 128 prompts \cite{wang2024inferaligner}, 200 sentences per concept \cite{dong24}, 256 strings \cite{zhang24}, 50 percent of the benchmark\cite{zhang2024truthx, wang24steering}, 500 questions from AdvBench \cite{feng24}, 1000 strings \cite{postmus24}, 1092 pairs of data \cite{chen24}, 150 contrastive pairs \cite{wang24}, subsets of benchmarks \cite{liu2024aligning}, 3000 to 8000 hateful samples \cite{garg24}, and up to 10,000 contrastive examples \cite{lee24}. PaCE, uses a large dictionary of over 40,000 concepts to create the final steering vector \cite{luo24}. SAFT uses k components ranging from 1 to 32 \cite{choi24}. 

It's easy to be misled into thinking a representation truly encodes a desired attribute when it is, in fact, merely coincidentally linked. Imagine you're trying to reduce the toxicity of a model. You identify a representation that's correlated with non-toxic responses. However, it turns out that this vector is also linked to the avoidance of certain controversial topics, even if those topics aren't inherently toxic. Intervening on the representation could inadvertently lead the model to censor valuable discussions, thus resulting in a degradation of helpfulness. The choice of stimuli and their number is critical to making sure the right representation has been identified.

Steering vectors can inadvertently encode spurious factors from the training data rather than the intended behavior. This means a steering vector might not correspond to the desired concept and may only be effective due to unintended biases. For example, a dataset used to create a "truthful" steering vector containing mostly simple sentences might only steer the model towards generating simple sentences rather than genuinely truthful ones. Spurious correlations might lead to steering vectors that amplify irrelevant features, causing inconsistent activation magnitudes across different inputs. This is problematic, as the theory behind what constitutes an optimal intervention is still imprecise. With so many different parameters for identifying representations, it becomes difficult to understand the optimal choice of stimuli and compare between methods.

\subsubsection{Token Choice} Token choice refers to which token(s) and their corresponding activations are used to construct the steering vectors. Selecting the right tokens from which to extract representations is crucial, as it directly influences the effectiveness and preciseness of the steering vector. The main differences in implementation are whether the steering vectors are calculated using a single token's activation, or across the activations of multiple tokens. Single-token approaches require formatting the contrastive inputs such that the inputs differ only by a single term. For example, methods such as CAA and ActAdd focus on the activations for yes/no. \cite{panickssery24, mattturner24}. In PACE, the activations from the last token of the generated output are used. This approach allows for the desired behavior to be isolated in a singular token, thus minimizing the complexity and noise that would have otherwise been introduced in an open-ended generation settings. Common methods for choosing this token include binary choice tokens like yes/no \cite{panickssery24, mattturner24, jorgensen24, luo2024pace} the token preceding the model's prediction, or the last token in the models' output\cite{zou2023representation}. In contrast, multiple-token approaches take the differences in activations across multiple tokens as opposed to a single one. Some methods such as Mean-centering \cite{jorgensen2023improving}, Category-wise Steering Vectors \cite{bhattacharjee24}, or specific RepEng applications for more general representations use more token activations to compute \cite{zou2023representation}.  

\subsubsection{Intervention Strength}

After a steering vector has been created, it is usually multiplied by a certain value before intervention. If this value is higher than 1, it implies the effect would be higher than if the representation was directly added through for example prompt engineering. If the scaling factor (which controls the strength of the steering intervention) is too large, adding the steering vector can drastically alter the activation norms, violating the magnitude consistency property and potentially leading to non-sensical output. Conversely, if the scaling factor is too low to preserve the activation norms, the steering vector may not be strong enough to shift the activation towards the desired behavior, hindering editing performance. Stronger scaling can disrupt coherence, while weaker scaling may be ineffective. Intervention strength beyond the steering vector norm causes a quadratic decrease in helpfulness (otherwise known as fluency), showing additional steering comes at a performance \cite{wolf24}. This means that only small enough interventions yield efficient steering with minimal performance loss. Excessive intervention strength degrades model coherence, leading to nonsensical generations \cite{rutte24, mattturner24}. However, when applying the intervention on one layer only, interventions with large co-efficents (larger than 1) do not lead to a visible deterioration of performance \cite{panickssery24}. 
Different studies vary intervention sizes. Some use low values of 0.25 \cite{zou2023representation}. Some use high values from 2 to 4 \cite{cao2024nothing, zou24circuit} or perform grid search to find the optimal intervention strength \cite{lee24}. Adaptive vector application strategies calculate intervention strength at inference \cite{stoher24, tlaie24, scalena24, kong24, wang24steering}. Interventions can also be negative. That is, a particular representation can be multiplied by a negative number to achieve its opposite. Intervention strength is one of the parameters affecting total effect on steering but cannot be analyzed in isolation- steering type and layer choices have been shown to affect optimal intervention strength. 

\subsubsection{Separation Method}


Once the neural activity has been identified, a classifier is required in order to map the models' internal activations to a specific concept or representation. The canonical RepE paper introduces both supervised linear models, including linear probing and differences between cluster means, as well as unsupervised linear models, such as PCA and K-means, for mapping internal states to concepts. However, benchmarking those methods against each other shows that PCA-based techniques are not optimal for identifying the target representation \cite{im2025unified}. Inference-time Interventions (ITI) fits a binary classifier onto the model's internal activations using the training, highlighting differences between generation accuracy (the model's output) and probe accuracy (the classifier output). PACE employs the same PCA-based classifer as the original representation engineering paper. Other alternatives to PCA have been explored, such as the Average Gradient Outer Product (AGOP)~\cite{beaglehole25}.

\subsubsection{Layers}

Some interventions aim to add the contrast vector to one layer only \cite{panickssery24, cao24steering, bhattacharjee24, cao24steering, jorgensen24, chalnev24,ferrando24halluc}. Those are always middle layers, usually 15 or 16 for 7-8B parameter models \cite{jorgensen24, rahn24,wu24, choi24, zou24circuit}. Some interventions add to a subset of layers \cite{postmus24, zhang24, liu2024aligning, zhao24, zhang2024truthx, wang2024inferaligner, pham24, zou2023representation}. Sometimes interventions happen on consecutive layers (e.g. 12-24 \cite{wang2024inferaligner} ) and sometimes not (e.g. 23-25 layers and 29-31 \cite{zhao24}). The rationale for choosing these layers is either inspired by other studies or comes from the best performing probing layers \cite{pham24, zhang2024truthx}. No studies intervene on early layers exclusively. This is confirmed by the fact that probing classification performance improves with deeper layers \cite{alain16}. This middle layer intervention strategy is supported by hyperparameter brute force search that shows intervention on one specific layer is approximately the same as on all of them, and that an intervention on the middle layers is the most effective \cite{subramani22}. However, other studies find hyperparameter searches leading to different optimal layers for each representation and model \cite{stolfo24}. 

Probing shows that in early layers, non-linear probing brings best effects, but the linearity of representations increases in later layers \cite{canby2024measuring}. Perhaps this can explain why interventions relying on linearity of representations perform well in middle layers. In fact, sparse probing explains that the middle layers often contain more dedicated neurons for higher-level linguistic and contextual features compared to earlier layers \cite{gurnee2023finding}. 

Probing shows that LLMs encode context knowledge in a layer-dependent manner, with upper layers retaining more contextual information, while lower layers primarily focus on entity-related features \cite{ju2024large}. Intermediate layers in large-language models (LLMs) consistently provide more informative representations for downstream tasks than final layers \cite{skean2024does, belrose2023eliciting}. Deeper layers are shown to contain more factual knowledge \cite{manigrasso2024probing, bronzini2024unveiling, troshin2022probing, aghazadeh2022metaphors, hernandez2022ast}.

Studies dedicated to layers show theory explaining these phenomena. Skean et al. \cite{skean2024does} judges representation quality at various layers through metrics such as prompt entropy, curvature, and augmentation invariance, showing that models undergo significant information compression at intermediate depths. Comparisons between Transformers and State Space Models (SSMs) show that Transformers exhibit more pronounced representational transformations, while SSMs maintain stable intermediate-layer representations. Training progression analysis indicates that the largest improvements in representation quality occur in intermediate layers, reinforcing their role in effective feature extraction. Intermediate layers exhibit bi-modal entropy distributions, where activations cluster into two distinct groups: one with high entropy, encoding diverse and context-sensitive information, and another with low entropy, representing more stable and deterministic knowledge. This suggests that intermediate layers naturally separate different types of information processing, which has significant implications for representation engineering. Activation steering and concept vector extraction can be more effectively applied to the low-entropy subset for stable modifications, while high-entropy activations may be leveraged for tasks requiring greater contextual flexibility. The structured separation also explains why mid-layer interventions, such as Inference-Time Intervention (ITI) and Sparse Autoencoder Steering, are more effective than modifications at early or final layers. Additionally, targeted entropy-aware interventions could enhance model alignment by suppressing high-entropy features to reduce hallucinations or amplifying low-entropy representations to improve factual consistency. Recognizing bimodal entropy as an emergent property of LLMs provides a framework for designing more precise and efficient steering methods that align with the model’s intrinsic representational structure.

However, while several studies support this finding, these have been conducted mostly on small-scale models (under 10B parameters). Also, function vectors work better in early layers \cite{todd2023function}, suggesting optimal layer depends on the intervention type. As the models get more complex, it becomes critical to conduct studies of larger size models to identify the optimal intervention layers using the new metrics. Therefore, as a fallback option, combinatorial testing, involving exhaustive brute force hyperparameter search through all existing combinations of layers is employed to identify the most effective combination of hyperparameters to detect and edit a representation. This is important because several studies show concepts do not remain stable across the network layers, implying constant interventions across all layers should not be applied \cite{nicolson24explain}. Experiments are primarily conducted on Llama-7B and 8B, with only some studies testing on larger models like Mixtral-8x7b \cite{rahn24} or 13B parameter models \cite{cao2024nothing, wang2024inferaligner}. Nonetheless, some studies apply interventions to all layers \cite{wolf24, mattturner24, luo24, dong24, wang24}. 

\subsubsection{Steering Type}

In the process of altering the output response style, the intervention may cause issues due to too strong of an intervention and its duplicate effect from intervening on multiple tokens. For instance, one can consider the intervention switching the token probability to be more positive.  If the first word of the response is changed from a normal to a positive one, the intervention effect for the second and any subsequent tokens will be magnified, as the probabilities of subsequent tokens depend on the previous sequence of tokens. This problem led to the emergence of several steering types. The \textbf{Initial} steering type only applies the intervention on the first token of the response \cite{todd2023function, mattturner24, leong23}. The \textbf{Constant} steering type maintains the same steering on each token of the response \cite{liu2023icv, cao24steering, panickssery24, luo24}. The \textbf{Diminishing} steering type starts with a full-strength response but reduces the intervention strength for each subsequent token \cite{scalena24}. 

\subsection{Multiple Model Methods}

Instead of operating on only one model, some techniques utilize multiple models to transplant concepts between models, improve the performance of the model after alignment (KTS \cite{stickland24}), and improve model safety (InferAligner \cite{wang2024inferaligner}, CONTRANS \cite{dong24} or KTCR \cite{garg24}). By utilizing the fact that different models have different infrastructures, one can leverage differing levels of safety between them to improve one of them or keep the performance stable after an intervention.
KL-Then-Steer (KTS) is a method that fine-tunes models to minimize Kullback–Leibler (KL) divergence between steered and unsteered outputs, reducing the impact of steering on the general capabilities of the model \cite{stickland24}. KTS prevents 44 percent of jailbreak attacks on Llama-2-Chat-7B while preserving general capabilities, outperforming system prompts and LoRA fine-tuning in maintaining model performance. The method generalizes to other interventions. The results suggest that KTS enables reliable control of the model without performance side-effects. In a similar fashion, InferAligner modifies model activations at inference time using safety steering vectors (SSVs) extracted from safety-aligned models \cite{wang2024inferaligner}. These vectors shift activations in specific representational directions to guide responses away from harmful outputs while preserving performance on downstream tasks. Steering vectors are transferable across models, so alignment-related representations exist in structured and reusable forms. Activation shifts effectively enforce harmlessness constraints without degrading the model’s general capabilities. Also using multiple models for safety, CONTRANS shows how alignment from smaller, aligned language models can be transferred to a larger, unaligned model using concept transplantation \cite{dong24}. CONTRANS extracts concept vectors from a source model using a small set of positive and negative examples, reformulates them through affine transformation to match the target model’s feature space, and then integrates them into the target model’s residual stream. Experimental results show that this method successfully transfers alignment concepts like truthfulness and fairness across different model sizes and families. CONTRANS achieves competitive performance compared to instruction tuning and reinforcement learning-based alignment while avoiding the high computational costs associated with those methods. Also attempting to work with safety concepts, KTCR shows a new method for implicit hate speech detection by refining conceptual representations through knowledge transfer \cite{garg24}. KTCR employs a teacher-student approach, where a pre-trained teacher model guides a student model in learning subtle hate speech patterns using concept activation vectors and prototype alignment. Unlike traditional data augmentation methods, KTCR not only introduces new implicit hate examples but also systematically refines how models internalize and distinguish such patterns. Therefore, this method is particularly powerful for detecting and editing concepts the definition of which changes over time.  

\subsection{Fine-tuning Alternatives}

Fine-tuning alternatives take a different approach from traditional weight-based fine-tuning by working directly in the model’s hidden representation space. Instead of updating large portions of the network’s weights, these methods target internal activations or add lightweight adapters, reducing both the risk of unintended side effects and the computational cost. By enhanced steerability and lightweight computational cost, representation engineering helps to change the performance and style of the model, hence fulfilling a similar role as fine-tuning but at a greatly decreased cost.

\subsubsection{TruthX}

Truthfulness in large-language models (LLMs) is not solely determined by their knowledge but also by how their internal representations influence response generation \cite{zhang2024truthx}. Some erroneous activations within these representations can cause hallucinations even when the correct knowledge is present. By identifying and modifying the truthful space of these representations, LLMs can be guided to produce more truthful responses without sacrificing their generative capabilities. The effectiveness of such interventions, like TruthX, depends on mapping representations into distinct semantic and truthful latent spaces, then applying contrastive learning to separate truthfulness from other linguistic properties. Editing these representations in the truthful space rather than directly manipulating outputs or attention patterns enables more consistent and controllable truthfulness enhancement across different LLMs.

\subsubsection{SAFT}

Safety-Aware Fine-Tuning (SAFT) aims to mitigate the risks posed by harmful data in the fine-tuning process of large-language models (LLMs), similar to the approach in Rosati et al. \cite{rosatirepresentation} and Choi et al \cite{choi24}. The framework employs a scoring function that detects and removes potentially harmful samples by analyzing the subspace information of LLM embeddings. Empirical results demonstrate that SAFT can reduce harmfulness by up to 27.8 percent across various models and contamination levels without significantly compromising model helpfulness. Traditionally, the paper examines the representation space of LLMs and how harmful samples can be identified through singular vector decomposition, contributing to a broader understanding of representation engineering in LLM fine-tuning.

\subsubsection{ReFT}

Representation Finetuning (ReFT) provides a parameter-efficient alternative to traditional weight-based fine-tuning by modifying hidden representations in large-language models (LLMs) instead of updating model weights \cite{wu24}. This method utilizes Low-rank Linear Subspace ReFT (LoReFT), which apply structured interventions in learned activation subspaces to improve model adaptation while using 15x–65x fewer parameters than LoRA. LoReFT consistently outperforms state-of-the-art parameter-efficient fine-tuning (PEFT) methods on common sense reasoning, instruction-following, and natural language understanding tasks. Unlike prior PEFTs, which modify a subset of weights, LoReFT operates entirely within the model’s latent space, preserving pre-trained knowledge while steering model behavior toward task-specific objectives. ReFT is therefore a cheaper model adaptation method that reduces the overhead of full fine-tuning.

\subsubsection{RAHF}

Representation Alignment from Human Feedback (RAHF) is a representation engineering alternative to RLHF \cite{liu2024aligning}. RAHF can extract and manipulate activity patterns humans value positively, aligning model outputs with human preferences faster and cheaper than traditional fine-tuning. By leveraging contrastive learning and activation-based modifications, RAHF provides a scalable way to adjust LLM behavior without requiring extensive additional training or reward modeling.

\subsubsection{LoRRA}
\cite{zou2023representation} uses LoRRA (Low-Rank Representation Adaptation), a lightweight fine-tuning method that integrates low-rank adapters into attention weights and optimize them using a representation-based loss function, such as the Contrast Vector. These adapters, referred to as controllers, are trained during model adaptation and later merged into the model, ensuring no additional computational cost during inference.

\subsection{Sparse Autoencoders-based approaches}

Sparse autoencoder-based approaches use models that learn compact, focused representations of hidden activations to identify the signals most relevant to steering language model behavior. This section provides an overview of methods using inspiration from both mechanistic interpretability and representation engineering to combine sparse autoencoders with activation steering. By training sparse autoencoders to capture key features, these methods can identify areas where the model’s stored knowledge conflicts with its current context or where specific semantic cues are encoded. These methods show how sparse representations can improve both the interpretability and precision of activation-steering interventions.

\subsubsection{SPARE}

Sparse Auto-Encoder-based Representation Engineering (SPARE) identifies signals of knowledge conflict in the residual stream of mid-layers, allowing precise detection of instances where contextual and parametric knowledge disagree \cite{zhao24}. SPARE extracts functional features from pre-trained sparse autoencoders (SAEs) that govern knowledge selection, enabling targeted intervention to prioritize either memory-based or context-based knowledge. Experimental results on open-domain question-answering tasks show that SPARE improves knowledge selection accuracy by 10 percent over existing representation engineering methods and 15 percent over contrastive decoding. 

\subsubsection{SAE-TS}

Chalnev et al. \cite{chalnev24} present a method to refine steering vectors to better control the behavior of language models while minimizing unintended effects through SAE-Targeted Steering (SAE-TS), which utilizes sparse autoencoders (SAEs) to measure the causal impact of steering vectors. Through this, more granular model control is achieved. The approach differs from existing methods by directly optimizing steering interventions to align with specific SAE features, avoiding the unpredictability of contrastive activation addition (CAA) and direct SAE feature steering. The paper compares SAE-TS with other steering methods across various tasks, demonstrating its superior ability to maintain coherence while effectively steering model behavior. In addition, the paper develops an interactive tool, EffectVis, to visualize feature effects and improve interpretability, contributing to ongoing research in mechanistic interpretability and representation engineering.

\subsubsection{Self-knowledge Sparse Autoencoders}

Ferrando et al. \cite{ferrando24halluc} examine the ability of LLM to recognize known and unknown entities. Using sparse autoencoders, it identifies specific directions in the model's representation space that encode self-knowledge, checking whether the model can recall factual information about an entity. These learned directions can causally influence model behavior, such as inducing refusals for unknown entities or prompting hallucinations when recognition is  manipulated. The findings suggest that the fine-tuning of the chat model repurposes existing entity recognition mechanisms rather than creating entirely new refusal strategies. Furthermore, the study highlights how these entity recognition directions regulate attention mechanisms, affecting factual recall and uncertainty expression.

\subsection{Other Approaches}

Due to its relatively recent emergence, representation engineering has evolved to multiple new applications. This section outlines key use cases to present methods for applying representation engineering to new applications.

\subsubsection{ARE}

Adversarial Representation Engineering (ARE) enables targeted model editing by leveraging representation based discriminators to refine internal activations \cite{zhang24}. This is useful for detecting jailbreaking attempts by monitoring for unusual activation patterns and counteracting harmful requests by injecting safety steering vectors to guide models towards safe outputs. Unlike previous methods that rely on direct representation vector manipulation, ARE employs adversarial learning between a generator (the LLM) and a discriminator, iteratively refining activation patterns for alignment and jailbreak tasks. Experiments show that ARE can reduce refusal rates on harmful prompts from 20 percent to under 1 percent in LLaMA-2 while also enhancing truthfulness in TruthfulQA benchmarks. The method, however, has also been shown to have bidirectional editing capabilities, enabling alignment reinforcement, but also its removal.

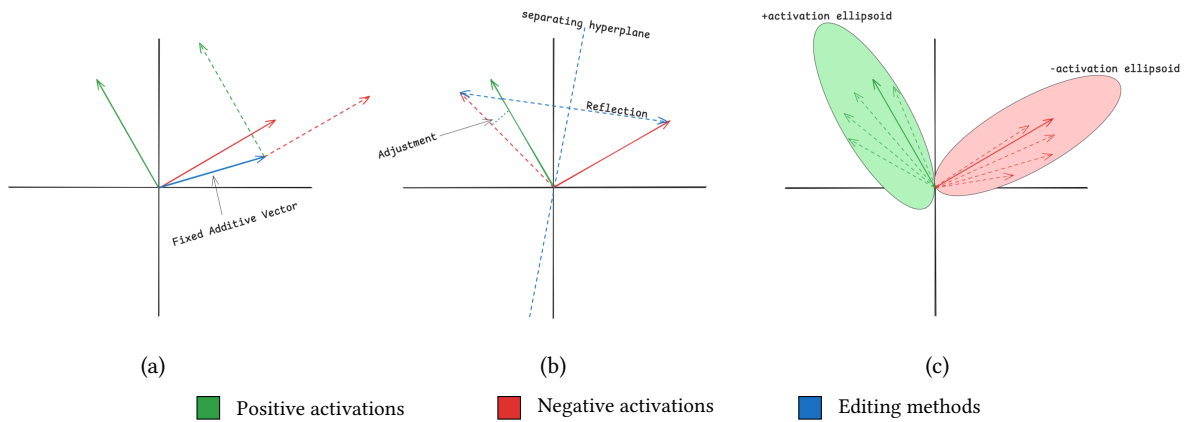
\begin{figure}[h]
    \centering
    \begin{tikzpicture}
        \node[inner sep=0pt] (image) at (0,0) {\includegraphics[width=\textwidth]{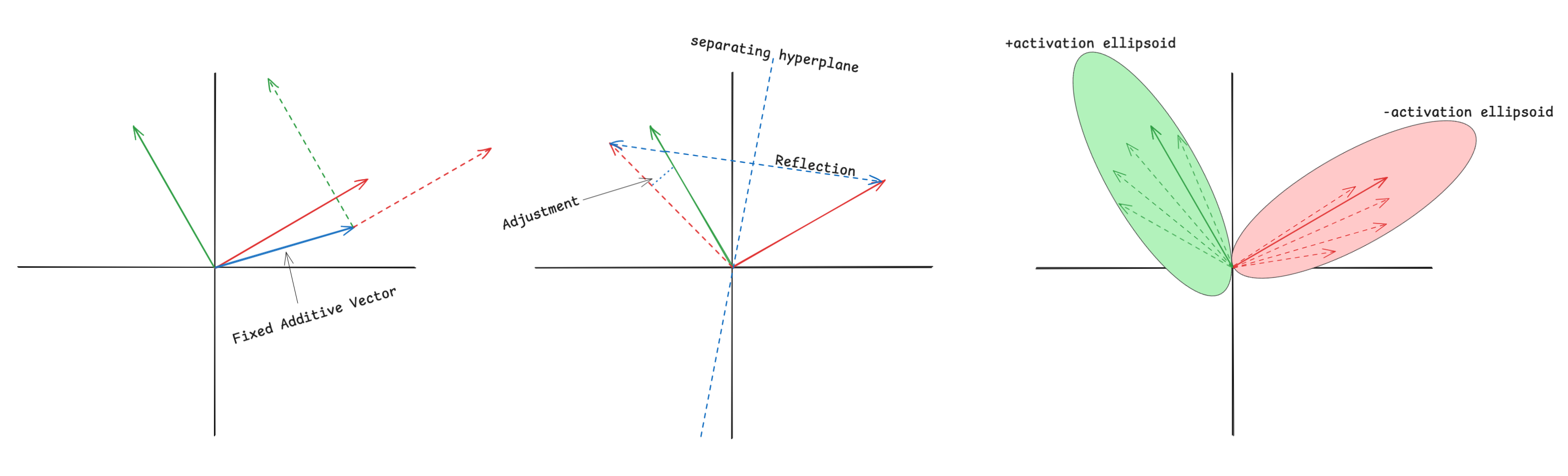}};
        \node[anchor=north, font=\small] at ([xshift=60pt]image.south west) {(a)};  
        \node[anchor=north, font=\small] at ([xshift=-15pt]image.south) {(b)}; 
        \node[anchor=north, font=\small] at ([xshift=-96pt]image.south east) {(c)};  
    \end{tikzpicture}
    \begin{tikzpicture}
  \node[draw, fill=constrast_steering_green, minimum size=0.3cm] (box1) at (-4, 0) {};
  \node[font=\footnotesize,right=0.1cm of box1] {Positive activations};
  
  \node[draw, fill=constrast_steering_red, minimum size=0.3cm] (box2) at (0, 0) {};
  \node[font=\footnotesize,right=0.1cm of box2] {Negative activations};
  
  \node[draw, fill=constrast_steering_blue, minimum size=0.3cm] (box3) at (4, 0) {};
  \node[font=\footnotesize,right=0.1cm of box3] {Editing methods};
  \end{tikzpicture}
    \caption{Geometric shift comparison of a few RepE techniques within the activation space. (a) represents additive steering \cite{panickssery24}, which translates activation vectors by a fixed steering vector, (b) depicts Householder Pseudo-Rotation (HPR) \cite{pham24}, which first reflects negative activations through a learned separating hyperplane and then adjusts the reflection to attain 90\degree 
 (this is an approximated rotation transformation), (c) depicts conceptor steering \cite{postmus24}, which bounds all activation vectors by softly projecting them onto a target ellipsoid.}

    \label{fig:comparison}
\end{figure}

\subsubsection{Conceptors}
Unlike traditional steering methods that rely on single additive steering vectors, Conceptors represent sets of activation vectors as ellipsoidal regions \cite{postmus24}. This allows for more structured manipulation of activations than a singular additive vector. Conceptors create a conceptor matrix from cached activations, which is then applied as a soft projection at inference time. Experiments on GPT-J and GPT-NeoX demonstrate that Conceptors outperform additive steering across multiple function-execution tasks, including antonym retrieval and language translation. Boolean operations on Conceptors, such as AND and OR, can be performed to combine multiple steering objectives more effectively than additive vector combination. 

\subsubsection{Circuit Breakers}

Circuit Breakers modify internal representations in large-language models (LLMs) to block harmful outputs at inference time, providing a more reliable safeguard than traditional refusal training \cite{zou24circuit}. The method, called Representation Rerouting (RR), redirects harmful activation patterns to predefined refusal states, preventing models from completing dangerous responses while preserving general capabilities. Circuit breakers can also effectively defend against adversarial image samples \cite{rando24}. Unlike adversarial training, which targets specific attack methods, RR operates on a model’s latent space, making it robust against unseen attacks, including embedding-space manipulations and multimodal adversarial inputs. Experiments on LLaMA-3 and Mistral-7B show that RR reduces harmful output rates by over 90 percent under diverse adversarial conditions while maintaining performance on standard benchmarks. 

\subsubsection{LEGEND}

\textbf{L}everaging r\textbf{E}presentation en\textbf{G}ineering to annotate prefer\textbf{EN}ce \textbf{D}atasets (LEGEND) automatically annotates safety margins in preference datasets using representation engineering \cite{feng24}. LEGEND identifies a "safety direction" within an LLM's embedding space by computing the difference between embeddings of harmful and harmless responses to increase safety with no need for extra training. As a purely inference time-method, it is more computationally efficient than the alternative margin annotation approaches. Experiments demonstrate that LEGEND improves reward modeling accuracy and enhances harmless alignment for LLMs, achieving comparable or superior results to ensemble-based methods while reducing annotation time significantly.

\subsection{Open source applications}

Representation engineering methods are also helpful for resolving issues related to open-source models. Representation Encoding Fingerprints (REEF) helps identify whether a suspect language model (LLM) is derived from a victim model by analyzing feature representations rather than weights \cite{zhang24reef}. Unlike weight-based fingerprinting, which is vulnerable to fine-tuning, pruning, and model merging, REEF computes the Centered Kernel Alignment (CKA) similarity between model representations on the same input samples, providing intellectual property protection. Representation Noising (RepNoise) is a defense mechanism designed to prevent harmful fine-tuning attacks (HFAs) on large-language models by removing information about harmful representations from model activations before an attacker gains access \cite{rosatirepresentation}. Unlike traditional safety guardrails that can be circumvented through fine-tuning, RepNoise alters intermediate activations such that recovering harmful behaviors through additional training becomes significantly more difficult. 

\subsection{Benefits of Representation Engineering}

Representation engineering seeks to answer questions critical to the development of model AI. In the process of representation learning, models generalize information to perform well on unseen tasks, transforming specific data points into useful heuristics for solving other tasks \cite{bengio13}. A large part of deep learning is dedicated to modifying training methods so that they make those generalizations better, for example through regularization\cite{kukačka2017regularizationdeeplearningtaxonomy}. In that, it aims to resolve the same fundamental problems as reinforcement learning from human feedback\cite{kaufmann2024surveyreinforcementlearninghuman}, direct preference optimization, fine-tuning, mechanistic interpretability and other approaches. 

\paragraph{Semantic Meaning}

Representation engineering facilitates the extraction of model representations and assigns them semantic meanings, enabling both the inspection of a model's reasoning and interpretation by human operators. By elucidating internal model concepts, this approach offers several beneficial applications. For instance, it allows for the establishment of precise guardrails to detect and mitigate harmful or biased patterns \cite{kotek2023gender}. Additionally, it enhances model optimization by enabling users to specify desired activation pattern shifts using natural language, as demonstrated by the Activation Addition technique \cite{mattturner24}. Furthermore, representation engineering aids in comparing models to select the most suitable one for specific use cases by providing semantically rich characterizations of model properties \cite{radfordclip}, thereby reducing reliance on generic benchmarks. 

\paragraph{Interpretability and Explainability}
Representation engineering provides a direct window into the internal reasoning of models by analyzing and manipulating their learned representations. This approach allows for the identification and mitigation of biases, harmful behaviors, and unintended model outputs at their root. By inspecting and assigning semantic meaning to these representations, it becomes possible to detect and address issues such as model "breaking free" or acting against user intentions \cite{li24, zhang24}. This level of transparency not only enhances trust in model behavior, but also enables the development of more robust safeguards against undesirable outcomes. Unlike post-hoc explainability methods, representation engineering operates at the foundational level of model reasoning, offering a more systematic and actionable approach to understanding and controlling model behavior.

\paragraph{Alignment and Personalization}
Representation engineering enables fine-grained control over model behavior by shifting representations to align with user-specific values and goals. Unlike traditional alignment methods, which often impose broad social or ethical values determined by model developers, representation engineering allows for individualized customization. Users can amplify or suppress specific representations to reflect their unique preferences, such as prioritizing honesty, creativity, or domain-specific expertise \cite{zou2023representation, weij24, zhang2024personalization, cao24steering}. This flexibility ensures that models can be tailored to diverse use cases without being constrained by a one-size-fits-all approach. By decoupling alignment from predefined social norms, representation engineering empowers users to define and enforce their own value systems, making models more adaptable and personally relevant.

\paragraph{Increasing Performance}
Representation engineering improves model performance by enabling precise interpretations as well as modifications to activation patterns. By identifying and amplifying representations associated with desired behaviors—such as accuracy, coherence, or task-specific expertise—users can optimize models for specific objectives \cite{zou2023representation, cao2024nothing, hendel2023context, bronzini2024unveiling, rahn24, stoher24, scalena24, wu24, zhao24, postmus24, liu2024aligning}. This approach contrasts with traditional fine-tuning, which often requires extensive retraining and may inadvertently degrade performance on unrelated tasks. Representation engineering allows for targeted adjustments that improve performance without compromising generalization. For example, in a medical diagnosis task, representations linked to "clinical accuracy" can be strengthened, leading to more reliable outputs. This precision makes representation engineering a powerful tool for optimizing models in a cost-effective and efficient manner.

\paragraph{Inference Time Control}
Representation engineering operates at inference time, eliminating the need for costly and invasive retraining. Traditional training methods are resource-intensive, unpredictable, and often require extensive setup. In contrast, inference-time control allows users to experiment with different configurations on-the-fly, adjusting representations to achieve desired outcomes without modifying the underlying model \cite{rimsky2023steering, mattturner24}. This approach reduces latency, lowers costs, and makes advanced AI capabilities more accessible. By using pretrained models, users can achieve precise control over model behavior without the overhead of additional training that often introduces high costs and takes a long time to implement. 

\section{Comparison with Other Approaches}

\begin{table}[h]
    \centering
    \begin{tabular}{lcccc}
        \toprule
        \textbf{Method} & \textbf{Steer Strength} & \textbf{Semantic Meaning} & \textbf{Computational Cost} & \textbf{Inference Time} \\
        \midrule
        Representation Engineering & \green & \green & \green & \green \\
        Sparse Autoencoders & \green & \green & \yellow & \yellow \\
        Prompt Engineering & \red & \yellow & \green & \yellow \\
        Soft Prompts & \yellow & \red & \green & \green \\
        Fine-Tuning & \yellow & \red & \red & \red \\
        Mechanistic Interpretability & \yellow & \green & \red & \yellow \\
        \bottomrule
    \end{tabular}
    \caption{Comparison of Representation Engineering with Other Approaches.}
    
    \green\: High \quad
    \yellow\: Medium \quad
    \red\: Low 
\end{table}

\subsection{Sparse Autoencoders}

Sparse autoencoders are neural networks designed to learn efficient feature representations by enforcing sparsity in the hidden layer activations, meaning only a small subset of neurons are active at a given time \cite{berahmand2024autoencoders}. Sparse autoencoders usually feature the addition of a sparsity constraint, such as L1 regularization or Kullback-Leibler divergence, to approximate most activations to zero while preserving important features \cite{zhai2018autoencoder}. Sparse autoencoders learn feature representations by enforcing sparsity constraints within the model activations, while representation engineering involves manually designing or modifying features to improve model performance. Unlike representation engineering, which relies on human intuition and domain knowledge, sparse autoencoders automatically discover structured, high-dimensional representations through unsupervised learning. Sparse autoencoders have been extensively applied to identify circuits of casual importance in large-language models \cite{marks24sparse}.

However, sparse autoencoders are not good for identifying steering vectors \cite{mayne24}. Steering vectors are out-of-distribution for SAEs, as they have significantly lower L2 norms than typical model activations, causing the SAE encoder bias to dominate the decomposition, and (2) SAEs enforce non-negative reconstruction coefficients, preventing them from capturing meaningful negative projections in feature directions. Furthermore, sparse autoencoders are much more computationally demanding and provide more difficult to interpret semantic meaning.

\subsection{Prompt Engineering}

Prompt engineering refers to a wide range of methods that aim to use more precise combinations of tokens on the input and system prompts to steer the model towards particular results \cite{sahoo2024systematic}. Prompt engineering focuses on crafting inputs to influence a model’s outputs, whereas representation engineering modifies how data is encoded within the model itself \cite{marvin2023prompt}. Unlike representation engineering, which alters feature representations inside the model, prompt engineering works externally by optimizing input structures without changing the model’s learned parameters. Therefore, the actual effect tokens have on the model internal representations is less interpretable and the steering less powerful.  

\paragraph{Soft Prompts}

Soft prompts are trainable input embeddings that guide frozen language models without modifying their parameters, exhibiting unique properties distinct from natural language prompts \cite{bailey2023soft}. Unlike discrete prompts, soft prompts occupy separate regions in embedding space, show heightened sensitivity to directional perturbations, and enable parameter-efficient adaptation for downstream tasks, as demonstrated in prefix-tuning \cite{li2021prefix}, prompt tuning \cite{lester2021power}, and P-Tuning \cite{liu2024gpt}. Studies show that soft prompts can enhance generalization in low-resource settings \cite{li2021prefix}, improve stability through continuous embeddings \cite{liu2024gpt}, and support novel applications like unlearning without weight updates \cite{bhaila2024soft}. However, their interpretability remains limited, and their susceptibility to adversarial manipulation raises security concerns \cite{bailey2023soft}. New, more efficent soft prompting methods, such as InfoPrompt, use information-theoretic objectives to optimize prompt initialization and task relevance, accelerating convergence and outperforming conventional tuning methods \cite{wu2024infoprompt}.

Soft prompting involves adding a learnable vector embeddings to the input sequence of a frozen language model, while representation engineering directly manipulates the internal activation patterns of the model without modifying its weights. In soft prompting, only the parameters of the added prompt vectors are updated during training, typically using a small prompt encoder, whereas representation engineering involves a more complex intervention across multiple layers of the model’s hidden states. The implementation of soft prompting requires minimal changes to the model architecture, often just extending the input processing pipeline, but representation engineering requires white-box access to the model to fully optimise the representation.

\subsection{Fine-tuning}


Fine-tuning refers to the process of shifting the weights of a pre-trained model by training it on a task-specific dataset \cite{dodge2020fine}. Fine-tuning utilizes transfer learning to shift the model from a general purpose one to a more specific one, adding skills the model previously did not have or domain expertise that was not represented in the training data \cite{zheng2023learn}. Like representation engineering, fine-tuning seeks to adapt a pre-trained model to a specific task or domain. Unlike representation engineering, the changes made through fine-tuning do not have a semantic meaning. Fine-tuning can be computationally expensive and may require significant resources, especially for large models, while representation engineering methods like Representation Fine-Tuning (ReFT) can be more parameter-efficient and less resource-intensive \cite{wu24}. In fact, using representation engineering to fine-tune representations more efficently leads to modifying less than 1 percent of the overall model representations to achieve comparable shifts in performance to traditional fine-tuning and are further optimized by solutions such as Low-rank Linear Subspace ReFT (LoReFT), a representation-engineering based alternative to LoRa adapters. 

When compared with fine-tuning directly, activation steering can additionally steer the model beyond what is possible with fine-tuning \cite{panickssery24, mattturner24}. However, combining these together leads to unpredictible interactions \cite{panickssery24}. In comparison to fine-tuning, representation engineering can allow to change intervention at steering time with "online steering", which fine-tuning does not allow \cite{mattturner24}. 

\subsection{Mechanistic Interpretability}

Mechanistic interpretability analyzes the contributions of specific parts of the network to model outcomes \cite{bereska24, rai2024practical}. In that, it looks at granular parts of the network and attempts to extract information on the interactions within the network. Mechanistic interpretablity uses neurons and circuits as fundamental units of analysis, hence focusing on identifying particular low-level mechanisms through which a model undertakes a decision \cite{zhao24survey, zhao24exppersp}. 

However, it is demanding to identify and locate circuits within a network. There is no guarantee that all parts of the network can be interpreted as circuits \cite{zou2023representation, hernandez24}. The research that publishes successful examples of circuits often presents a misleading picture of how difficult it is to assign an actual representation to the circuit, as negative results are not published \cite{bereska24}. Furthermore, given the use of interpretability techniques in training, worries about the mechanistic interpretability techniques actually intensifying the adversarial pressure against interpretability increase. This means that as new techniques to detect power-seeking or otherwise deceptive behaviour of LLMs are used, the deceptive model will make it harder to find such interpretable circuits but not change its intentions and behavior. 

Mechanistic interpretability has the same goal as representation engineering. Mechanistic interpretability and representation engineering can be seen as opposing approaches: one takes the bottom-up perspective of analyzing building blocks of the network to understand its overall performance while the other adopts a top-down approach, using global representations to extract meaning \cite{zou2023representation}. However, in practice, the methods are hard to clearly distinguish. Activation patching for example, can be seen as a specific case of a very granular representation engineering intervention. In comparison with existing mechanistic interpretability methods like automated circuit discovery \cite{conmy2023towards}, attribution patching \cite{syed2023attribution},  causal scrubbing \cite{brinkmann2024mechanistic} or sparse autoencoder-based neuron interpretation \cite{cunningham2023sparse}, representation engineering  tends to produce more immediate and controllable changes in model behavior in a more lightweight manner than SA-based mechanistic interpretability approaches and enables direct manipulation of high-level concepts without requiring a full mechanistic understanding of the underlying circuits. 

\subsection{Combining Representation Engineering With Other Methods}

Using activation steering does not prevent the users from using alternatives. Representation engineering have also been successfully used as a part of a "modular LLM" architecture, selectively adding representation engineering interventions combined with other interventions to steer the model \cite{xiao24config}. Different compatibility metrics have been developed, showing the order of these interventions matters for their overall effectiveness \cite{kolbeinsson24}. 

\section{Evaluation}

Understanding whether the representation has been sucessfully detected and steered is a critical component of building a reliable representation engineering framework. The lack of a standardized approach hinders the effectiveness and adoption of representation engineering. In particular, open-ended generation evaluation is important. Interventions that perform well in multiple-choice formats can fail in open-ended scenarios. CAA \cite{panickssery24} has been shown to successfully steer the model in a multiple-choice setup but failed in an open-ended context using the same prompts and interventions \cite{pres24}. In understand the effectivness of a particular intervention, model size and architecture are important. Representation engineering works better for larger models \cite{bhattacharjee24}. This is potentially because larger models build more detectable linear representations \cite{gurnee23}. Different models have differing representations, but steering vectors are transferable over architectures and model sizes \cite{cao24steering}. 

\subsection{Measures of steering}

\subsubsection{Task Accuracy}

Most evaluations report simply the changes in performance on a particular dataset. For hallucination detection, this is usually TruthfulQA \cite{zhang24, chen24, dong24, stickland24, zou2023representation}, with one paper evaluating on NQSwap and Macnoise \cite{zhao24}. Evaluations on TruthfulQA, however, often make use only of the multiple-choice version of the benchmark \cite{cao24steering, pham24, li2024inference}. Similarly, MMLU is also a multiple choice benchmark. This is potentially problematic, as it does not allow to examine the degradation of performance as the result of steering.  

For safety interventions, methods are evaluated using different benchmarks, including BeaverTails \cite{bhattacharjee24, choi24, scalena24}, AdvBench \cite{cao24steering, cao2024nothing}, HarmBench \cite{zou24circuit}, ToxiGen \cite{dong24, wang24} or other more niche benchmarks. Similarly, when evaluating general model performance, MMLU \cite{panickssery24, luo24, chen24}, Alpaca \cite{bhattacharjee24, scalena24} or CatQA \cite{bhattacharjee24, mattturner24} are used, with several other benchmarks such as ARC-Easy or AQuA used in one paper each. 

The fact that outputs change or remain stable in a given direction does not necessarily indicate the presence of a meaningful linear representation, as changes might result from unrelated geometric artifacts in the model’s representation space \cite{park24}. Instead of relying solely on benchmark outputs, token probabilities in a given context should be evaluated to determine whether interventions in the representation space effectively steer model behavior \cite{park24}. Concept representations should be tested for their alignment with counterfactual token pairs to validate their linear structure, hence showing that differences between words expressing the same conceptual change point in a consistent direction. For example, CAA \cite{panickssery24} aimed at inducing myopic behavior showed nearly equal probabilities for both myopic and non-myopic tokens, suggesting that difference in outputs between steered and unsteered output depends on sampling randomness rather than a fundamental shift in model behavior \cite{pres24}. Rather than using arbitrary inner products, the causal inner product should be estimated and assessed for its ability to enforce orthogonality between causally separable concepts. This allows to show that changes in one concept do not unintentionally affect another, degrading fluency on other tasks. In addition, representation reading can be used for validation to verify that the computed concept representations can predict target attributes in a logit-linear way. Also, shifts in token probabilities should be analysed while increasing the intervention strength and checking if that leads to the expected emergence of higher probability of the target concept in next-token predictions \cite{park24}.

Alternatives to benchmark performance include the Kullback–Leibler divergence (KL) of the
model’s next-token prediction distribution post- versus pre-intervention \cite{li2024inference},  reduction in perplexity \cite{mattturner24} or latent separation scores, logit difference measurements, and attention score changes \cite{ferrando24halluc}. 

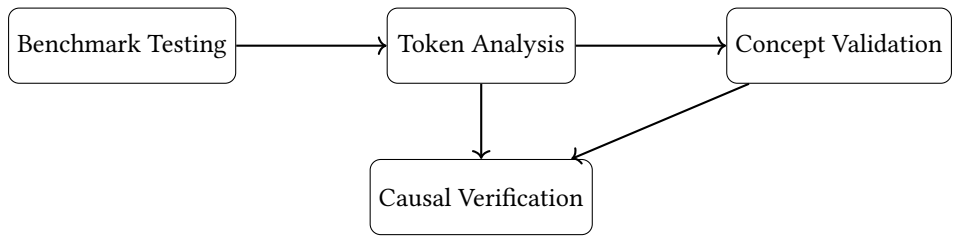
\begin{figure}[H]
\centering
\begin{tikzpicture}[
    block/.style={rectangle, draw, rounded corners, minimum width=2.5cm, minimum height=1cm},
    arrow/.style={->, thick}
]
    \node[block] (step1) {Benchmark Testing};
    \node[block, right=2cm of step1] (step2) {Token Analysis};
    \node[block, right=2cm of step2] (step3) {Concept Validation};
    \node[block, below=1cm of step2] (step4) {Causal Verification};
    
    \draw[arrow] (step1) -- (step2);
    \draw[arrow] (step2) -- (step3);
    \draw[arrow] (step2) -- (step4);
    \draw[arrow] (step3) -- (step4);
\end{tikzpicture}
\caption{Recommended Evaluation Pipeline}
\label{fig:pipeline}
\end{figure}

\subsection{Measures of fluency}

Steering vectors can reduce harmful output frequency but often degrade performance on benign inputs \cite{stickland24}. While interventions can often improve alignment metrics, such as decreasing the rate of successful jailbreaks by 30\% for specific models, it's also possible that these same interventions, without proper consideration, might reduce the model's accuracy on standard question-answering or other general capabilities \cite{weij24,wolf24,zhao24}, sometimes by as much as 15-20\% on MT-Bench \cite{stickland24, zou24circuit}. This is often seen where targeting too strongly a desired behavior will also break other model behaviors.  In particular, for RepE \cite{zou2023representation} interventions beyond the steering vector norm lead to rapid degradation of performance \cite{wolf24}. Large steering vectors guarantee intervention success but reduce model outputs to near-random guessing \cite{wolf24}. Therefore, each intervention must be precisely calibrated to avoid excessive degradation of general capabilities. The lack of theoretical validation for the choice of hyperparameters and standardized methods of evaluation exacerbates this problem. Therefore, evaluations often consider both the effect of steering on outputs while measuring linguistic fluency \cite{li2024inference, stolfo24}. Several metrics used to track this include perplexity \cite{mattturner24}, perplexity-normalized effect size (PNES) \cite{rutte24}, BLEU \cite{pham24, subramani22}, ROUGE-L \cite{choi24} and  generation entropy or the weighted average of bi- and tri-gram entropies \cite{brumley24}. Nonetheless, lack of loss of fluency on a particular representation intervention does not imply the model will perform maintain coherence on other tasks that will have different representation parameters assigned by the search of optimal layer or other model parameters. 

\section{Open Problems in Representation Engineering}


Despite significant progress in representation engineering, several challenges remain unresolved. This section examines key open problems identified in recent research, their implications for the field's development, and potential directions for future work.

\subsection{Standardization}

\subsubsection{Lack of Standardized Evaluation}

Many activation engineering techniques employ different methods to evaluate the effectiveness of steering. Evaluations often do not assess out-of-distribution inputs, limiting the generalizability of the findings \cite{bereska24}. \cite{pres24} found that an intervention exhibited corrigible behavior in a multiple-choice format but failed to do so in an open-ended generation setting, highlighting the importance of measuring steering interventions on tasks similar to the eventual use-case. It critiques existing evaluation methods for activation steering in large-language models and proposes a standardized evaluation pipeline to measure intervention effectiveness. It identifies four key missing properties in current assessments: \emph{a)} evaluations should be conducted in open-ended generation contexts, \emph{b)} account for model likelihoods, \emph{c)} allow for standardized cross-behavior comparisons, and \emph{d)} include baseline comparisons. Using this framework, the study evaluates two activation steering methods: Contrastive Activation Addition (CAA) and Inference-Time Intervention (ITI) on behaviors such as truthfulness, corrigibility, and myopia. The results reveal that the effectiveness of the intervention is highly behavior-dependent, and ITI successfully increases the likelihood of true responses, while CAA is more effective in reducing hallucinated outputs than in promoting desired behaviors. The paper also finds that multiple choice evaluations overstate intervention success compared to open-ended settings.

\subsubsection{Systematizing Parameter Interventions}

As outlined in this paper, the exact structure of each intervention differs significantly. The lack of a standardized approach to representation engineering can make it difficult to compare and contrast different methods, especially in the context of the overall intervention strength. The lack of standardized parameters, including stimuli number, token choice (encoder vs. decoder, specific tokens), separation method, intervention strength (constant, diminishing, initial), and target layer(s), makes it difficult to compare results across different studies and hinders the development of generalizable best practices.  This variability, coupled with the context-dependent nature of knowledge encoding in LLMs, where representations shift across layers and tokens, creates a complex search space for optimal RepE intervention and representation evaluation.  Consequently, computationally expensive brute-force hyperparameter searches need to be implemented to find the optimal intervention for a given use case. The abscence of a detailed theoretical framework on optimal choices of these parameters makes practical implementation of RepE a challenge and requires further investigation. 

\subsubsection{Lack of Theoretical Grounding}

We lack a solid theoretical framework to explain why interventions work, whether they should be theoretical, and often we see that they work best within a given range of coefficients. For example, it's difficult to predict in advance which layers are most receptive to steering, with some studies finding that middle layers work best \cite{mattturner24,rahn24,zhao24}, while others find that the efficacy varies based on the specific task. Furthermore, steerability appears to be a property of the dataset rather than the model, since different architectures exhibit similar patterns of reliability and failure \cite{tan24}. The extent to which each kind of vector can steer effectively is limited by whether the portion of latent space from which they are extracted actually contains a discoverable representation of the information needed to execute the desired behavior \cite{brumley24}. 

Crucially, the effectiveness of steering vectors has shed insights on the linear representation hypothesis (LRH). LRH claims a stronger global linearity in feature space. These studies show there are representations for which the linear hypothesis is true, but also some for which it likely does not hold \cite{ark24, mallen2024elicitinglatentknowledgequirky, marks24, tan24}. Whether the representations are linear likely also depends on model size. Since interventions are tested on small models of similar sizes, it is important to create a generalizable theory that unifies interventions, representations, parameters, and models in a tractable framework that does not rely on empirical search to find optimal parameters but instead develops theoretical explanations for their optimal values.

\subsubsection{Challenges in Generalization}

Steering vectors show high variance in effectiveness, with some datasets exhibiting nearly 50 percent anti-steerable examples, where the intervention produces the opposite of the intended effect. Generalization performance is strongly correlated with the similarity between the source and target prompts, suggesting that steering vectors work best when applied to behaviors the model already tends to produce. Additionally, it can be difficult to predict the behavior of steering vectors. In some cases, steering interventions do not produce interpretable changes in model behavior other than model degradation. Also, the out-of-distribution generalization for many behaviors is not perfect or entirely predictable, making it difficult to precisely control the model \cite{chalnev24}. 

This unpredictability often leads to an empirical, trial-and-error approach, which hinders efficiency and limits our ability to generalize interventions and understand the underlying mechanisms. The sign (direction) and overall effectiveness of steering interventions can vary widely across and within different concepts. This highlights a key limitation: the extent to which steering works is dependent on whether the desired information is already embedded in the model’s latent space. The fact that different architectures exhibit similar patterns of reliability and failure further complicates the challenge of generalizing interventions across models \cite{tan24}.

\subsection{Steering Multimodal Models}

Although there is extensive literature on steering large-language models, the rise of multimodal models has not led to the widespread successful application of RepE techniques in modalities like video and images. Control vectors can also be used for improving motion control and forecasting \cite{tas24}. Multimodal models incorporate tools from multiple modalities to create a cohesive tool for generating outputs across computer vision, text and video. Representations like safety and truthfulness require designing interventions that hold across different modalities, since for jailbreaks can be performed across different modalities. However, research on representation learning shows it is likely that multimodality will introduce additional problems. Additional research is needed to create reliable multimodal representation engineering. In particular, these problems are likely to occur: 

\subsubsection{Negative Transfer \& Disentanglement of Features}
Expanding the space of interventions to multimodal :
\emph{a)} Interventions aimed at enhancing one modality may inadvertently degrade the quality or semantic integrity of another. This misalignment occurs because feature spaces across modalities do not necessarily share common transformation characteristics.
\emph{b)} The resulting representations may lose their semantic validity, particularly when interventions fail to account for the distinct ways in which meaning is encoded across different modalities.
\emph{c)} The high dimensionality of multimodal feature spaces complicates the identification of meaningful and salient directions for intervention.
For example, consider a video depicting culinary preparation. A RepE intervention designed to \emph{increase happiness} might unintentionally amplify the visual prominence of certain ingredients or generate artifacts in video frames that bear little relation to the subject's emotional state. This phenomenon stems from the challenge of accurately modeling intricate cross-modal dependencies, where the relationships between static elements in images and dynamic actions in video sequences are often subtle and context-dependent.

\subsubsection{Tokenization and Encoding Bottleneck}
RepE approaches require a continuous and differentiable latent space. However, modern multimodal architectures often employ non-differentiable encoding mechanisms, such as tokenization or latent vector quantization, to integrate information across modalities \cite{vaswani2017attention}. These discontinuous transformations disrupt gradient propagation.
Rando et al. \cite{rando24} identify this challenge, noting that to enable continuous end-to-end optimization in tokenization-based models, a \emph{tokenizer shortcut} must be implemented. Although this modification facilitates gradient-based adversarial attacks, it introduces additional complications, notably the tendency toward overconfident token predictions. This trade-off highlights a fundamental drawback in the design of robust representation systems: enabling gradient flow for optimization purposes may create vulnerabilities that can be exploited by adversarial methods.
\begin{algorithm}
\caption{Multimodal Representation with Tokenization Challenge}
\label{alg:tokenization}
\begin{algorithmic}[1]
\Require Text input $x_{text}$, Image input $x_{image}$
\Require Encoders $E_{text}$, $E_{image}$
\Require Tokenizer $T$, Quantizer $Q$

\Function{ProcessMultimodal}{$x_{text}, x_{image}$}
    \State // Initial encoding
    \State $z_{text} \gets E_{text}(x_{text})$ \Comment{Continuous text embedding}
    \State $z_{image} \gets E_{image}(x_{image})$ \Comment{Continuous image embedding}
    
    \State // Non-differentiable transformations
    \State $t_{text} \gets T(z_{text})$ \Comment{Discrete tokens}
    \State $t_{image} \gets Q(z_{image})$ \Comment{Quantized vectors}
    
    \State // Gradient flow blocked here
    \State \textbf{Problem:} $\nabla_{z_{text}}t_{text} = 0$ and $\nabla_{z_{image}}t_{image} = 0$
    
    \State // Tokenizer shortcut (Rando et al.)
    \State $t'_{text} \gets T_{\text{continuous}}(z_{text})$ \Comment{Differentiable approximation}
    
    \Return $t'_{text}, t_{image}$
\EndFunction
\end{algorithmic}
\end{algorithm}

\subsubsection{Semantic Meaning for Latent Space Directions}
Decomposing \emph{What does a specific vector addition mean visually?} by interpreting the meaning of directional manipulations in the latent spaces of visual models is difficult. Vector manipulations in the latent space of video generation models may induce perceptible changes to the generated content without affecting the targeted semantic attribute. 
This interpretability gap can be partially addressed by deploying auxiliary models designed for cross-modal alignment. For example, CLIP \cite{radfordclip} facilitates the measurement of cross-modal similarities by projecting textual descriptions into the same embedding space as visual content.

\subsubsection{Non-linear Interactions}

The \emph{linear representation hypothesis} suggests that high-level concepts are represented linearly in intermediate LLM activations \cite{scalena24}. If features are not linearly separable or are interdependent, simple vector addition might not isolate and modify a concept without affecting others. If features are non-linear, simply adding a steering vector might disproportionately affect certain dimensions, disrupting the original activation magnitude balance \cite{tlaie24}. For representation reading, this means that linear probes may not capture the complexity of the representations.The detected representations can be misleading or fail to detect certain features that have a non-linear character or trigger only through interactions with other features \cite{bereska24}. Some methods attempt to measure and create non-linear interactions, such as CAST \cite{lee24} or MiMiC \cite{singh24} have been developed, but creating more advanced methods with empirical validation should be a key priority for creating effective interventions, especially for smaller models.

\subsection{Ethical Challenges}

\paragraph{Dual Use}
Representation engineering, while powerful for steering language model behavior, presents a dual use challenge, as it can be employed for both beneficial and potentially harmful purposes. Any beneficial intervention can be easily reversed, and steering vectors can be used to effectively undo existing safety guardrails embedded in LLMs \cite{zou2023representation}. Through RepE, potentially harmful biases can be introduced into the model activation space, shifting the model behavior in an undetectable fashion \cite{meng2023locating}. Because of that, it becomes critical to devise and engineer methods to reliably detect representations that have been tampered with and devise methods for embedding a beneficial representation in an irreversible way in the model so that it cannot be reversed.

\paragraph{Bias Amplification, Propagation, and Suppression}
LLMs often inherit biases from their training data. Representation engineering could inadvertently amplify these biases or introduce new ones, leading to unfair or discriminatory results \cite{luo24,tlaie24,zhao24survey}. RepE could be used to suppress certain viewpoints or censor content, raising concerns about freedom of expression. For example, the overly aggressive use of RepE to remove \emph{undesirable} concepts could lead to the suppression of legitimate discourse \cite{cao2024nothing}. Therefore, it is important to create solutions that steer the model without exaggerated effects.

\section{Conclusion}

This survey has presented a comprehensive overview of representation engineering in LLMs, highlighting both current achievements and future challenges. As the field continues to evolve, addressing the identified research challenges will be crucial for advancing our understanding and control of more powerful models.

\begin{acks}
We gratefully acknowledge the support of AI Safety Camp and the Feuer Scholarship at the University of Warwick.
\end{acks}

\bibliographystyle{ACM-Reference-Format}
\bibliography{bibliography}

\end{document}